\documentclass[conference]{IEEEtran}
\IEEEoverridecommandlockouts

\usepackage{cite}
\usepackage{amsmath,amssymb,amsfonts}
\usepackage{algorithmic}
\usepackage{graphicx}
\usepackage{textcomp}
\usepackage{xcolor}
\usepackage[bookmarksnumbered,unicode]{hyperref}
\hypersetup{hidelinks}
\hypersetup{pdflang={en},pdfdisplaydoctitle}
\usepackage[font=small,format=plain,labelfont=bf,textfont=bf]{caption}
\usepackage[skip=0cm,list=true]{subcaption}
\usepackage{xpatch}

\usepackage[frozencache,cachedir=minted-cache]{minted}
\usepackage{xspace}

\usepackage{scalerel}
\usepackage{tikz}
\usetikzlibrary{svg.path}

\definecolor{orcidlogocol}{HTML}{A6CE39}
\tikzset{
  orcidlogo/.pic={
    \fill[orcidlogocol] svg{M256,128c0,70.7-57.3,128-128,128C57.3,256,0,198.7,0,128C0,57.3,57.3,0,128,0C198.7,0,256,57.3,256,128z};
    \fill[white] svg{M86.3,186.2H70.9V79.1h15.4v48.4V186.2z}
                 svg{M108.9,79.1h41.6c39.6,0,57,28.3,57,53.6c0,27.5-21.5,53.6-56.8,53.6h-41.8V79.1z M124.3,172.4h24.5c34.9,0,42.9-26.5,42.9-39.7c0-21.5-13.7-39.7-43.7-39.7h-23.7V172.4z}
                 svg{M88.7,56.8c0,5.5-4.5,10.1-10.1,10.1c-5.6,0-10.1-4.6-10.1-10.1c0-5.6,4.5-10.1,10.1-10.1C84.2,46.7,88.7,51.3,88.7,56.8z};
  }
}

\newcommand\orcidicon[1]{\href{https://orcid.org/#1}{\mbox{\scalerel*{
\begin{tikzpicture}[yscale=-1,transform shape]
\pic{orcidlogo};
\end{tikzpicture}
}{|}}}}

\definecolor{cf09021}{RGB}{240,144,33}
\definecolor{cefb61d}{RGB}{239,182,29}
\definecolor{ccccd2b}{RGB}{204,205,43}
\definecolor{c07a6aa}{RGB}{7,166,170}

\tikzset{
  kaustlogo/.pic={
  \path[fill=cf09021] (2.8502, 1.6591).. controls (2.9163, 1.7583) and (2.9494, 1.8812) .. (2.9447, 2.0041).. controls (2.9447, 2.0797) and (2.9353, 2.1554) .. (2.9116, 2.231).. controls (2.836, 2.1648) and (2.7935, 2.0703) .. (2.7935, 1.971).. controls (2.7982, 1.8623) and (2.8171, 1.7583) .. (2.8502, 1.6591);

  \path[fill=cf09021] (3.6632, 2.4673).. controls (3.6868, 2.3444) and (3.6962, 2.2215) .. (3.6962, 2.0986).. controls (3.6962, 1.6401) and (3.6064, 1.314) .. (3.4694, 1.0777).. controls (3.493, 1.073) and (3.5166, 1.073) .. (3.5403, 1.073).. controls (4.1831, 1.073) and (4.3202, 1.6685) .. (4.3202, 1.9285).. controls (4.3202, 2.3491) and (3.8948, 2.4579) .. (3.6632, 2.4673);

  \path[fill=cf09021] (3.0156, 1.9994).. controls (3.0203, 1.8481) and (2.9731, 1.7016) .. (2.8785, 1.5834).. controls (2.9022, 1.522) and (2.9353, 1.4653) .. (2.9683, 1.4085).. controls (3.096, 1.5551) and (3.1574, 1.763) .. (3.1574, 2.0325).. controls (3.1574, 2.1412) and (3.1385, 2.2499) .. (3.1054, 2.3491).. controls (3.0581, 2.3255) and (3.0156, 2.3019) .. (2.9731, 2.2688).. controls (3.0014, 2.1884) and (3.0156, 2.0939) .. (3.0156, 1.9994);

  \path[fill=cf09021] (3.2378, 2.0325).. controls (3.2378, 1.7016) and (3.148, 1.4842) .. (3.0156, 1.3424).. controls (3.0487, 1.2998) and (3.0912, 1.262) .. (3.1338, 1.2242).. controls (3.2756, 1.4038) and (3.3701, 1.6685) .. (3.3701, 2.0655).. controls (3.3701, 2.1884) and (3.3559, 2.3066) .. (3.3228, 2.4248).. controls (3.2756, 2.4106) and (3.2283, 2.3964) .. (3.1858, 2.3822).. controls (3.2189, 2.2735) and (3.2425, 2.1554) .. (3.2378, 2.0325);

  \path[fill=cf09021] (3.4694, 2.0655).. controls (3.4694, 1.6401) and (3.3654, 1.3566) .. (3.2141, 1.1628).. controls (3.2567, 1.1344) and (3.3039, 1.1155) .. (3.3512, 1.1013).. controls (3.4883, 1.3235) and (3.5781, 1.6449) .. (3.5781, 2.0986).. controls (3.5781, 2.2215) and (3.5639, 2.3444) .. (3.5403, 2.4673).. controls (3.5025, 2.4626) and (3.4599, 2.4579) .. (3.4126, 2.4484).. controls (3.4505, 2.3255) and (3.4694, 2.1979) .. (3.4694, 2.0655);

  \path[fill=cefb61d] (2.1837, 3.0298).. controls (2.5382, 3.0298) and (2.8313, 2.8927) .. (3.0203, 2.6706).. controls (3.0581, 2.7226) and (3.0912, 2.7793) .. (3.1196, 2.836).. controls (2.9494, 3.0487) and (2.6895, 3.1905) .. (2.335, 3.1905).. controls (2.0466, 3.181) and (1.763, 3.129) .. (1.4889, 3.0298).. controls (1.4936, 2.992) and (1.5078, 2.9542) .. (1.5267, 2.9211).. controls (1.7394, 2.992) and (1.9616, 3.0298) .. (2.1837, 3.0298);

  \path[fill=cefb61d] (3.2141, 3.3039).. controls (3.1905, 3.6253) and (3.0203, 4.1547) .. (2.4295, 4.1547).. controls (1.971, 4.1547) and (1.6969, 3.7246) .. (1.5692, 3.3937).. controls (1.867, 3.4883) and (2.1743, 3.5403) .. (2.4815, 3.545).. controls (2.7462, 3.5544) and (3.0062, 3.4694) .. (3.2141, 3.3039);

  \path[fill=cefb61d] (1.6732, 2.7367).. controls (1.7914, 2.7651) and (1.9143, 2.7745) .. (2.0372, 2.7745).. controls (2.2357, 2.7745) and (2.6186, 2.7367) .. (2.8407, 2.4957).. controls (2.888, 2.5288) and (2.9305, 2.5618) .. (2.9683, 2.6044).. controls (2.7982, 2.8171) and (2.5193, 2.9494) .. (2.1837, 2.9494).. controls (1.9757, 2.9494) and (1.7678, 2.9163) .. (1.5645, 2.8502).. controls (1.5976, 2.8076) and (1.6354, 2.7698) .. (1.6732, 2.7367);

  \path[fill=cefb61d] (2.7793, 2.4626).. controls (2.5713, 2.6753) and (2.2215, 2.7084) .. (2.0419, 2.7084).. controls (1.9427, 2.7084) and (1.8434, 2.6989) .. (1.7489, 2.6847).. controls (1.9757, 2.5193) and (2.3019, 2.4011) .. (2.5146, 2.4011).. controls (2.5997, 2.3964) and (2.6942, 2.4153) .. (2.7793, 2.4626);

  \path[fill=cefb61d] (2.335, 3.2898).. controls (2.7036, 3.2898) and (2.9778, 3.148) .. (3.1621, 2.9353).. controls (3.1858, 3.0062) and (3.2047, 3.0771) .. (3.2141, 3.1527).. controls (3.0534, 3.3181) and (2.8171, 3.4316) .. (2.4815, 3.4316).. controls (2.1554, 3.4221) and (1.8292, 3.3607) .. (1.522, 3.2567).. controls (1.5125, 3.2189) and (1.5031, 3.1763) .. (1.4936, 3.1385).. controls (1.763, 3.233) and (2.0466, 3.285) .. (2.335, 3.2898);

  \path[fill=ccccd2b] (0.7894, 2.3964).. controls (0.7894, 2.6658) and (0.9595, 2.8596) .. (1.1958, 3.0014).. controls (1.1628, 3.0251) and (1.1249, 3.044) .. (1.0871, 3.0534).. controls (0.865, 2.9116) and (0.7137, 2.7178) .. (0.7137, 2.472).. controls (0.7137, 2.2499) and (0.7941, 1.9663) .. (0.9406, 1.6921).. controls (0.9642, 1.6921) and (0.9879, 1.6969) .. (1.0115, 1.7063).. controls (0.8697, 1.9474) and (0.7894, 2.2026) .. (0.7894, 2.3964);

  \path[fill=ccccd2b] (0.9784, 2.3208).. controls (0.9784, 2.5524) and (1.1297, 2.7178) .. (1.3329, 2.8313).. controls (1.3187, 2.8644) and (1.2951, 2.8974) .. (1.2715, 2.9258).. controls (1.0493, 2.8029) and (0.8886, 2.6327) .. (0.8886, 2.3917).. controls (0.8886, 2.2073) and (0.9642, 1.971) .. (1.0966, 1.7394).. controls (1.1202, 1.7583) and (1.1391, 1.7725) .. (1.158, 1.7961).. controls (1.0446, 1.9852) and (0.9831, 2.1743) .. (0.9784, 2.3208);

  \path[fill=ccccd2b] (0.9312, 3.0865).. controls (0.9028, 3.0912) and (0.8744, 3.0912) .. (0.8461, 3.0912).. controls (0.5767, 3.0912) and (0.0, 2.8691) .. (0.0, 2.4579).. controls (0.0, 2.0183) and (0.4868, 1.7489) .. (0.8035, 1.6969).. controls (0.6665, 1.9757) and (0.5908, 2.2546) .. (0.5908, 2.472).. controls (0.5956, 2.7273) and (0.7279, 2.9305) .. (0.9312, 3.0865);

  \path[fill=ccccd2b] (1.1628, 2.2499).. controls (1.1628, 2.4011) and (1.2431, 2.5335) .. (1.3849, 2.6233).. controls (1.3849, 2.6706) and (1.3755, 2.7178) .. (1.3613, 2.7604).. controls (1.1817, 2.6611) and (1.0588, 2.5193) .. (1.0588, 2.3255).. controls (1.0682, 2.1648) and (1.1202, 2.0088) .. (1.2053, 1.867).. controls (1.2242, 1.8954) and (1.2384, 1.9285) .. (1.2526, 1.9568).. controls (1.2006, 2.0466) and (1.1675, 2.1459) .. (1.1628, 2.2499);

  \path[fill=ccccd2b] (1.2337, 2.2499).. controls (1.2384, 2.1743) and (1.2573, 2.1034) .. (1.2904, 2.0372).. controls (1.3471, 2.1979) and (1.3802, 2.3681) .. (1.3896, 2.5382).. controls (1.2904, 2.472) and (1.2337, 2.3633) .. (1.2337, 2.2499);

  \path[fill=c07a6aa] (2.2593, 1.366).. controls (1.9663, 1.3707) and (1.6874, 1.4842) .. (1.4794, 1.6827).. controls (1.4416, 1.6685) and (1.4085, 1.6543) .. (1.3755, 1.6354).. controls (1.5976, 1.3755) and (1.9237, 1.1628) .. (2.3161, 1.1628).. controls (2.4815, 1.158) and (2.6422, 1.1911) .. (2.7887, 1.262).. controls (2.7651, 1.3282) and (2.732, 1.3896) .. (2.6895, 1.4416).. controls (2.5477, 1.3896) and (2.4059, 1.3613) .. (2.2593, 1.366);

  \path[fill=c07a6aa] (1.0021, 1.3755).. controls (0.865, 1.2384) and (0.7799, 1.0871) .. (0.7799, 0.9312).. controls (0.7799, 0.3356) and (1.7063, 0.0) .. (2.0655, 0.0).. controls (2.4909, 0.0) and (2.6895, 0.2694) .. (2.7745, 0.553).. controls (2.6564, 0.5247) and (2.5335, 0.5105) .. (2.4106, 0.5105).. controls (1.815, 0.5152) and (1.3235, 0.9075) .. (1.0021, 1.3755);

  \path[fill=c07a6aa] (1.919, 1.7772).. controls (1.7914, 1.7772) and (1.6685, 1.7536) .. (1.5503, 1.711).. controls (1.7441, 1.5362) and (1.9994, 1.4369) .. (2.2593, 1.4369).. controls (2.387, 1.4322) and (2.5146, 1.4558) .. (2.6327, 1.4983).. controls (2.4059, 1.73) and (2.0514, 1.7772) .. (1.919, 1.7772);

  \path[fill=c07a6aa] (2.3113, 1.0824).. controls (1.8907, 1.0824) and (1.5362, 1.3187) .. (1.2998, 1.5976).. controls (1.2715, 1.5834) and (1.2478, 1.5692) .. (1.2195, 1.5503).. controls (1.4794, 1.1958) and (1.8812, 0.8981) .. (2.3586, 0.8981).. controls (2.5193, 0.8933) and (2.68, 0.9264) .. (2.8265, 0.9879).. controls (2.8265, 1.054) and (2.8171, 1.1202) .. (2.8029, 1.1817).. controls (2.6469, 1.1155) and (2.4815, 1.0777) .. (2.3113, 1.0824);

  \path[fill=c07a6aa] (2.3586, 0.7988).. controls (1.8481, 0.7988) and (1.418, 1.1202) .. (1.1391, 1.4983).. controls (1.1202, 1.4842) and (1.1013, 1.4653) .. (1.0824, 1.4511).. controls (1.3849, 1.0068) and (1.8481, 0.6334) .. (2.4106, 0.6334).. controls (2.5429, 0.6334) and (2.6753, 0.6523) .. (2.8029, 0.6901).. controls (2.8171, 0.7563) and (2.8218, 0.8224) .. (2.8265, 0.8886).. controls (2.68, 0.8272) and (2.5193, 0.7988) .. (2.3586, 0.7988);
}
}

\newcommand\kausticon{\mbox{\scalerel*{
\begin{tikzpicture}[yscale=1,transform shape]
\pic{kaustlogo};
\end{tikzpicture}
}{|}}}

\definecolor{ce03128}{RGB}{224,49,40}

\tikzset{
  uljubljanalogo/.pic={
  \path[fill=ce03128] (0.2808, 0.766) -- (0.3497, 0.766) -- (0.3497, 0.8745).. controls (0.3497, 0.879) and (0.3487, 0.8835) .. (0.347, 0.8876).. controls (0.3452, 0.8917) and (0.3426, 0.8955) .. (0.3394, 0.8987).. controls (0.3362, 0.9018) and (0.3324, 0.9043) .. (0.3282, 0.906).. controls (0.324, 0.9076) and (0.3195, 0.9085) .. (0.315, 0.9084).. controls (0.3106, 0.9085) and (0.3061, 0.9076) .. (0.3019, 0.906).. controls (0.2978, 0.9043) and (0.294, 0.9018) .. (0.2908, 0.8986).. controls (0.2876, 0.8955) and (0.2851, 0.8917) .. (0.2834, 0.8876).. controls (0.2817, 0.8834) and (0.2808, 0.879) .. (0.2808, 0.8745) -- cycle;

  \path[fill=ce03128] (0.1834, 1.1201) -- (0.1145, 1.1201) -- (0.1145, 0.9776) -- (0.1834, 0.9776) -- cycle;

  \path[fill=ce03128] (1.181, 1.8881) -- (1.181, 1.8651) -- (1.181, 1.8568) -- (1.2268, 1.8568) -- (1.2268, 1.8881).. controls (1.2267, 1.9053) and (1.2212, 1.922) .. (1.2109, 1.9358).. controls (1.2006, 1.9496) and (1.1862, 1.9597) .. (1.1697, 1.9647) -- (1.1697, 2.3242) -- (1.124, 2.3242) -- (1.124, 1.9217) -- (1.1466, 1.9217).. controls (1.1557, 1.9218) and (1.1645, 1.9182) .. (1.171, 1.9118).. controls (1.1775, 1.9054) and (1.1812, 1.8968) .. (1.1813, 1.8876);

  \path[fill=ce03128] (0.682, 1.1201) -- (0.6131, 1.1201) -- (0.6131, 0.9776) -- (0.682, 0.9776) -- cycle;

  \path[fill=ce03128] (0.1147, 0.766) -- (0.1835, 0.766) -- (0.1835, 0.8745).. controls (0.1834, 0.879) and (0.1825, 0.8835) .. (0.1807, 0.8876).. controls (0.179, 0.8917) and (0.1764, 0.8955) .. (0.1732, 0.8987).. controls (0.17, 0.9018) and (0.1662, 0.9043) .. (0.162, 0.906).. controls (0.1578, 0.9076) and (0.1533, 0.9085) .. (0.1488, 0.9084).. controls (0.1443, 0.9085) and (0.1399, 0.9076) .. (0.1357, 0.906).. controls (0.1316, 0.9043) and (0.1278, 0.9018) .. (0.1246, 0.8986).. controls (0.1215, 0.8955) and (0.1189, 0.8917) .. (0.1172, 0.8876).. controls (0.1155, 0.8834) and (0.1147, 0.879) .. (0.1147, 0.8745) -- cycle;

  \path[fill=ce03128] (0.1834, 0.3731) -- (0.1145, 0.3731) -- (0.1145, 0.2701) -- (0.1834, 0.2701) -- cycle;

  \path[fill=ce03128] (0.6478, 0.909).. controls (0.6433, 0.9091) and (0.6388, 0.9082) .. (0.6346, 0.9066).. controls (0.6304, 0.9049) and (0.6266, 0.9024) .. (0.6234, 0.8993).. controls (0.6202, 0.8961) and (0.6176, 0.8923) .. (0.6158, 0.8882).. controls (0.6141, 0.884) and (0.6131, 0.8796) .. (0.6131, 0.8751) -- (0.6131, 0.766) -- (0.682, 0.766) -- (0.682, 0.8745).. controls (0.682, 0.879) and (0.6811, 0.8834) .. (0.6794, 0.8876).. controls (0.6777, 0.8917) and (0.6752, 0.8955) .. (0.672, 0.8986).. controls (0.6688, 0.9018) and (0.665, 0.9043) .. (0.6609, 0.906).. controls (0.6567, 0.9076) and (0.6522, 0.9085) .. (0.6478, 0.9084);

  \path[fill=ce03128] (0.5156, 1.1201) -- (0.4473, 1.1201) -- (0.4473, 0.9776) -- (0.5156, 0.9776) -- cycle;

  \path[fill=ce03128] (0.3498, 1.1201) -- (0.2809, 1.1201) -- (0.2809, 0.9776) -- (0.3498, 0.9776) -- cycle;

  \path[fill=ce03128] (0.4473, 0.766) -- (0.5156, 0.766) -- (0.5156, 0.8745).. controls (0.515, 0.8832) and (0.5112, 0.8913) .. (0.5049, 0.8972).. controls (0.4985, 0.9032) and (0.4902, 0.9065) .. (0.4815, 0.9065).. controls (0.4728, 0.9065) and (0.4644, 0.9032) .. (0.4581, 0.8972).. controls (0.4517, 0.8913) and (0.4479, 0.8832) .. (0.4473, 0.8745) -- cycle;

  \path[fill=ce03128] (0.0458, 1.1201) -- (0.0, 1.1201) -- (0.0, 0.1562) -- (0.6819, 0.1562) -- (0.6819, 0.2021) -- (0.0458, 0.2021) -- (0.0458, 0.6527) -- (0.6819, 0.6527) -- (0.6819, 0.6982) -- (0.0458, 0.6982) -- cycle;

  \path[fill=ce03128] (0.1834, 0.5841) -- (0.1145, 0.5841) -- (0.1145, 0.4415) -- (0.1834, 0.4415) -- cycle;

  \path[fill=ce03128] (0.9684, 1.7482) -- (1.3216, 1.7482) -- (1.3769, 1.5417) -- (1.4246, 1.5417) -- (1.357, 1.7937) -- (0.9346, 1.7937) -- (0.9331, 1.7942) -- (0.8653, 1.5421) -- (0.9131, 1.5421) -- cycle;

  \path[fill=ce03128] (0.682, 0.5841) -- (0.6131, 0.5841) -- (0.6131, 0.4415) -- (0.682, 0.4415) -- cycle;

  \path[fill=ce03128] (0.925, 0.4815) -- (0.925, 0.1141) -- (0.9708, 0.1141) -- (0.9708, 0.4815).. controls (0.9704, 0.5088) and (0.9594, 0.5349) .. (0.94, 0.554).. controls (0.9205, 0.5732) and (0.8944, 0.5839) .. (0.8671, 0.5839) -- (0.8649, 0.5839) -- (0.8649, 0.5383) -- (0.8671, 0.5383).. controls (0.8746, 0.5383) and (0.882, 0.5369) .. (0.889, 0.534).. controls (0.8959, 0.5312) and (0.9022, 0.527) .. (0.9076, 0.5217).. controls (0.9129, 0.5165) and (0.9171, 0.5102) .. (0.92, 0.5032).. controls (0.9229, 0.4963) and (0.9244, 0.4889) .. (0.9244, 0.4814);

  \path[fill=ce03128] (0.682, 0.3731) -- (0.6131, 0.3731) -- (0.6131, 0.2701) -- (0.682, 0.2701) -- cycle;

  \path[fill=ce03128] (0.3498, 0.3731) -- (0.2809, 0.3731) -- (0.2809, 0.2701) -- (0.3498, 0.2701) -- cycle;

  \path[fill=ce03128] (0.3498, 0.5841) -- (0.2809, 0.5841) -- (0.2809, 0.4415) -- (0.3498, 0.4415) -- cycle;

  \path[fill=ce03128] (0.5156, 0.3731) -- (0.4473, 0.3731) -- (0.4473, 0.2701) -- (0.5156, 0.2701) -- cycle;

  \path[fill=ce03128] (0.5156, 0.5841) -- (0.4473, 0.5841) -- (0.4473, 0.4415) -- (0.5156, 0.4415) -- cycle;

  \path[fill=ce03128] (1.3187, 0.766) -- (1.3876, 0.766) -- (1.3876, 0.8745).. controls (1.3875, 0.879) and (1.3866, 0.8835) .. (1.3848, 0.8876).. controls (1.3831, 0.8917) and (1.3805, 0.8955) .. (1.3773, 0.8987).. controls (1.3741, 0.9018) and (1.3703, 0.9043) .. (1.3661, 0.906).. controls (1.3619, 0.9076) and (1.3574, 0.9085) .. (1.3529, 0.9084).. controls (1.3484, 0.9085) and (1.344, 0.9076) .. (1.3398, 0.906).. controls (1.3356, 0.9043) and (1.3319, 0.9018) .. (1.3287, 0.8986).. controls (1.3255, 0.8955) and (1.323, 0.8917) .. (1.3212, 0.8876).. controls (1.3195, 0.8834) and (1.3187, 0.879) .. (1.3187, 0.8745) -- cycle;

  \path[fill=ce03128] (1.8457, 0.3731) -- (1.7774, 0.3731) -- (1.7774, 0.2701) -- (1.8457, 0.2701) -- cycle;

  \path[fill=ce03128] (1.6801, 0.5841) -- (1.6112, 0.5841) -- (1.6112, 0.4415) -- (1.6801, 0.4415) -- cycle;

  \path[fill=ce03128] (1.8457, 0.5841) -- (1.7774, 0.5841) -- (1.7774, 0.4415) -- (1.8457, 0.4415) -- cycle;

  \path[fill=ce03128] (1.8459, 0.6982) -- (1.6112, 0.6982) -- (1.6112, 0.6527) -- (1.8459, 0.6527) -- cycle;

  \path[fill=ce03128] (1.8459, 0.2021) -- (1.6112, 0.2021) -- (1.6112, 0.1562) -- (1.8459, 0.1562) -- cycle;

  \path[fill=ce03128] (1.6801, 0.3731) -- (1.6112, 0.3731) -- (1.6112, 0.2701) -- (1.6801, 0.2701) -- cycle;

  \path[fill=ce03128] (0.1621, 1.6112) -- (0.797, 1.6112) -- (0.797, 1.6567) -- (0.1269, 1.6567) -- (0.0004, 1.1886) -- (0.6818, 1.1886) -- (0.6818, 1.2347) -- (0.0606, 1.2347) -- cycle;

  \path[fill=ce03128] (1.8457, 1.1201) -- (1.7774, 1.1201) -- (1.7774, 0.9776) -- (1.8457, 0.9776) -- cycle;

  \path[fill=ce03128] (1.8459, 1.2347) -- (1.6112, 1.2347) -- (1.6112, 1.1888) -- (1.8459, 1.1888) -- cycle;

  \path[fill=ce03128] (1.6112, 0.766) -- (1.6801, 0.766) -- (1.6801, 0.8745).. controls (1.68, 0.879) and (1.6791, 0.8835) .. (1.6773, 0.8876).. controls (1.6756, 0.8917) and (1.673, 0.8955) .. (1.6698, 0.8987).. controls (1.6666, 0.9018) and (1.6627, 0.9043) .. (1.6586, 0.906).. controls (1.6544, 0.9076) and (1.6499, 0.9085) .. (1.6454, 0.9084).. controls (1.6409, 0.9085) and (1.6365, 0.9076) .. (1.6323, 0.906).. controls (1.6281, 0.9043) and (1.6244, 0.9018) .. (1.6212, 0.8986).. controls (1.618, 0.8955) and (1.6155, 0.8917) .. (1.6137, 0.8876).. controls (1.612, 0.8834) and (1.6111, 0.879) .. (1.6112, 0.8745) -- cycle;

  \path[fill=ce03128] (1.8459, 1.6567) -- (1.496, 1.6567) -- (1.496, 1.6112) -- (1.8459, 1.6112) -- cycle;

  \path[fill=ce03128] (1.496, 0.1305) -- (1.5423, 0.1562) -- (1.5423, 1.4512) -- (1.496, 1.4512) -- cycle;

  \path[fill=ce03128] (1.8116, 0.909).. controls (1.8072, 0.9091) and (1.8027, 0.9082) .. (1.7986, 0.9066).. controls (1.7944, 0.9049) and (1.7906, 0.9024) .. (1.7875, 0.8992).. controls (1.7843, 0.8961) and (1.7818, 0.8923) .. (1.7801, 0.8882).. controls (1.7784, 0.884) and (1.7775, 0.8796) .. (1.7776, 0.8751) -- (1.7776, 0.766) -- (1.8459, 0.766) -- (1.8459, 0.8745).. controls (1.8459, 0.879) and (1.845, 0.8834) .. (1.8433, 0.8876).. controls (1.8416, 0.8917) and (1.8391, 0.8955) .. (1.8359, 0.8986).. controls (1.8327, 0.9018) and (1.8289, 0.9043) .. (1.8247, 0.906).. controls (1.8206, 0.9076) and (1.8161, 0.9085) .. (1.8116, 0.9084);

  \path[fill=ce03128] (1.6801, 1.1201) -- (1.6112, 1.1201) -- (1.6112, 0.9776) -- (1.6801, 0.9776) -- cycle;

  \path[fill=ce03128] (1.3224, 0.4815) -- (1.3224, 0.1141) -- (1.3682, 0.1141) -- (1.3682, 0.4815).. controls (1.3682, 0.4891) and (1.3697, 0.4965) .. (1.3726, 0.5034).. controls (1.3755, 0.5103) and (1.3797, 0.5166) .. (1.385, 0.5219).. controls (1.3903, 0.5272) and (1.3967, 0.5314) .. (1.4036, 0.5342).. controls (1.4106, 0.537) and (1.418, 0.5385) .. (1.4255, 0.5384) -- (1.4277, 0.5384) -- (1.4277, 0.5841) -- (1.4255, 0.5841).. controls (1.3982, 0.5841) and (1.372, 0.5733) .. (1.3526, 0.5542).. controls (1.3332, 0.535) and (1.3221, 0.509) .. (1.3218, 0.4817);

  \path[fill=ce03128] (1.1811, 0.698) -- (1.1122, 0.698) -- (1.1122, 0.8296).. controls (1.1128, 0.8383) and (1.1166, 0.8465) .. (1.123, 0.8525).. controls (1.1294, 0.8585) and (1.1379, 0.8619) .. (1.1466, 0.8619).. controls (1.1554, 0.8619) and (1.1638, 0.8585) .. (1.1702, 0.8525).. controls (1.1766, 0.8465) and (1.1805, 0.8383) .. (1.1811, 0.8296) -- cycle(1.2267, 0.8296).. controls (1.2258, 0.8502) and (1.217, 0.8697) .. (1.2021, 0.884).. controls (1.1872, 0.8983) and (1.1673, 0.9063) .. (1.1466, 0.9063).. controls (1.126, 0.9063) and (1.1061, 0.8983) .. (1.0912, 0.884).. controls (1.0763, 0.8697) and (1.0674, 0.8502) .. (1.0665, 0.8296) -- (1.0665, 0.698) -- (0.8655, 0.698) -- (0.8655, 0.6525) -- (1.4285, 0.6525) -- (1.4285, 0.698) -- (1.2267, 0.698) -- cycle;

  \path[fill=ce03128] (1.3071, 1.4497) -- (1.1462, 1.5421) -- (1.1453, 1.5427) -- (0.9859, 1.4503) -- (0.9859, 1.3716) -- (0.8653, 1.3716) -- (0.8653, 1.3257) -- (1.0317, 1.3257) -- (1.0317, 1.4242) -- (1.1458, 1.4897) -- (1.2613, 1.4236) -- (1.2613, 1.3257) -- (1.4277, 1.3257) -- (1.4277, 1.3716) -- (1.3071, 1.3716) -- cycle;

  \path[fill=ce03128] (0.9056, 0.766) -- (0.9745, 0.766) -- (0.9745, 0.8745).. controls (0.9745, 0.879) and (0.9736, 0.8834) .. (0.9719, 0.8876).. controls (0.9702, 0.8917) and (0.9677, 0.8955) .. (0.9645, 0.8986).. controls (0.9613, 0.9018) and (0.9575, 0.9043) .. (0.9533, 0.906).. controls (0.9492, 0.9076) and (0.9447, 0.9085) .. (0.9402, 0.9084).. controls (0.9357, 0.9085) and (0.9313, 0.9076) .. (0.9271, 0.906).. controls (0.9229, 0.9043) and (0.9191, 0.9018) .. (0.9159, 0.8987).. controls (0.9126, 0.8955) and (0.9101, 0.8917) .. (0.9083, 0.8876).. controls (0.9065, 0.8835) and (0.9056, 0.879) .. (0.9056, 0.8745) -- cycle;

  \path[fill=ce03128] (1.2499, 0.1141) -- (1.2499, 0.4816).. controls (1.2475, 0.5073) and (1.2356, 0.5312) .. (1.2165, 0.5486).. controls (1.1974, 0.566) and (1.1725, 0.5757) .. (1.1466, 0.5757).. controls (1.1208, 0.5757) and (1.0959, 0.566) .. (1.0768, 0.5486).. controls (1.0577, 0.5312) and (1.0457, 0.5073) .. (1.0433, 0.4816) -- (1.0433, 0.1141) -- (1.089, 0.1141) -- (1.089, 0.4816).. controls (1.0897, 0.4963) and (1.0961, 0.5102) .. (1.1068, 0.5204).. controls (1.1176, 0.5306) and (1.1318, 0.5363) .. (1.1465, 0.5363).. controls (1.1613, 0.5363) and (1.1755, 0.5306) .. (1.1863, 0.5204).. controls (1.197, 0.5102) and (1.2034, 0.4963) .. (1.2041, 0.4816) -- (1.2041, 0.1141) -- cycle;

  \path[fill=ce03128] (0.7508, 0.1562) -- (0.797, 0.1305) -- (0.797, 1.4512) -- (0.7508, 1.4512) -- cycle;

  \path[fill=ce03128] (1.1927, 1.3651).. controls (1.1926, 1.3741) and (1.1898, 1.383) .. (1.1847, 1.3905).. controls (1.1796, 1.398) and (1.1724, 1.4039) .. (1.1639, 1.4073).. controls (1.1555, 1.4107) and (1.1463, 1.4116) .. (1.1374, 1.4098).. controls (1.1285, 1.4079) and (1.1203, 1.4035) .. (1.1139, 1.3971).. controls (1.1075, 1.3906) and (1.1031, 1.3824) .. (1.1014, 1.3735).. controls (1.0996, 1.3646) and (1.1005, 1.3554) .. (1.104, 1.347).. controls (1.1075, 1.3386) and (1.1134, 1.3314) .. (1.1209, 1.3263).. controls (1.1285, 1.3213) and (1.1373, 1.3186) .. (1.1464, 1.3185).. controls (1.1525, 1.3186) and (1.1585, 1.3198) .. (1.1642, 1.3221).. controls (1.1698, 1.3245) and (1.1749, 1.3279) .. (1.1792, 1.3322).. controls (1.1835, 1.3365) and (1.1869, 1.3417) .. (1.1892, 1.3473).. controls (1.1915, 1.3529) and (1.1927, 1.359) .. (1.1927, 1.3651) -- cycle;

  \path[fill=ce03128] (0.9745, 1.1201) -- (0.9056, 1.1201) -- (0.9056, 0.9776) -- (0.9745, 0.9776) -- cycle;

  \path[fill=ce03128] (1.1479, 1.28).. controls (1.1004, 1.2803) and (1.0543, 1.2642) .. (1.0172, 1.2345) -- (0.8655, 1.2345) -- (0.8655, 1.1886) -- (1.0342, 1.1886).. controls (1.0647, 1.2182) and (1.1054, 1.2347) .. (1.1478, 1.2347).. controls (1.1902, 1.2347) and (1.231, 1.2182) .. (1.2614, 1.1886) -- (1.4278, 1.1886) -- (1.4278, 1.2345) -- (1.2786, 1.2345).. controls (1.2414, 1.2641) and (1.1954, 1.2802) .. (1.1479, 1.28) -- cycle;

  \path[fill=ce03128] (1.2267, 0.9776) -- (1.2267, 1.0861).. controls (1.2258, 1.1067) and (1.217, 1.1262) .. (1.202, 1.1405).. controls (1.1871, 1.1548) and (1.1672, 1.1628) .. (1.1466, 1.1628).. controls (1.1259, 1.1628) and (1.1061, 1.1548) .. (1.0911, 1.1405).. controls (1.0762, 1.1262) and (1.0674, 1.1067) .. (1.0665, 1.0861) -- (1.0665, 0.9776) -- (1.1121, 0.9776) -- (1.1121, 1.0861).. controls (1.1127, 1.0948) and (1.1166, 1.103) .. (1.123, 1.109).. controls (1.1294, 1.115) and (1.1378, 1.1183) .. (1.1466, 1.1183).. controls (1.1553, 1.1183) and (1.1638, 1.115) .. (1.1702, 1.109).. controls (1.1766, 1.103) and (1.1805, 1.0948) .. (1.181, 1.0861) -- (1.181, 0.9776) -- cycle;

  \path[fill=ce03128] (1.3876, 1.1201) -- (1.3187, 1.1201) -- (1.3187, 0.9776) -- (1.3876, 0.9776) -- cycle;

  \path[fill=ce03128] (0.9257, 0.0455) -- (0.8653, 0.0797) -- (0.8653, 0.027) -- (0.9134, 0.0) -- (1.3796, 0.0) -- (1.4277, 0.027) -- (1.4277, 0.0797) -- (1.3673, 0.0455) -- cycle;

}
}

\newcommand\uljubljanaicon{\mbox{\scalerel*{
\begin{tikzpicture}[yscale=1,transform shape]
\pic{uljubljanalogo};
\end{tikzpicture}
}{|}}}

\newcommand{\iconbox}[1]{\raisebox{0pt}[0pt][0pt]{\textsuperscript{\footnotesize #1}}}

\newcommand{\colext}{\textsc{CoLExT}\xspace}

\newif\ifsubmission
\submissiontrue

\ifsubmission
\newcommand{\mcnote}[1]{}
\newcommand{\veljkonote}[1]{}
\newcommand{\janeznote}[1]{}
\newcommand{\amandionote}[1]{}

\else

\newcommand{\mcnote}[1]{\todo[color=orange!40,inline]{marco: #1}}
\newcommand{\veljkonote}[1]{\todo[color=blue!40,inline]{veljko: #1}}
\newcommand{\janeznote}[1]{\todo[color=green!40,inline]{janez: #1}}
\newcommand{\amandionote}[1]{\todo[color=magenta!40,inline]{amandio: #1}}
\fi

\newif\ifrevision
\revisiontrue
\ifrevision

\else

\fi

\AtBeginDocument{%
  \providecommand\BibTeX{{%
    \normalfont B\kern-0.5em{\scshape i\kern-0.25em b}\kern-0.8em\TeX}}}

\begin{document}

\title{Where is the Testbed for my\\ Federated Learning Research?}

\author{\IEEEauthorblockN{Janez Božič\iconbox{\uljubljanaicon}\iconbox{\kausticon}\IEEEauthorrefmark{1} \orcidicon{0009-0003-0115-5901},%
\IEEEcompsocitemizethanks{\IEEEcompsocthanksitem\IEEEauthorrefmark{1}Equal contribution. Work done in part while Janez Božič was interning at KAUST.}
Amândio R. Faustino\iconbox{\kausticon}\IEEEauthorrefmark{1} \orcidicon{0009-0009-6877-1999},
Boris Radovič\iconbox{\uljubljanaicon}\iconbox{\kausticon}\IEEEauthorrefmark{1} \orcidicon{0009-0008-4142-931X},
Marco Canini\iconbox{\kausticon} \orcidicon{0000-0002-5051-4283},
Veljko Pejović\iconbox{\uljubljanaicon} \orcidicon{0000-0002-9009-0024}}
\IEEEauthorblockA{\iconbox{\uljubljanaicon} University of Ljubljana$\qquad$\iconbox{\kausticon} KAUST}
}

\maketitle

\begin{abstract}

Progressing beyond centralized AI is of paramount importance, yet, distributed AI solutions, in particular various federated learning (FL) algorithms, are often not comprehensively assessed, which prevents the research community from identifying the most promising approaches and practitioners from being convinced that a certain solution is deployment-ready. The largest hurdle towards FL algorithm evaluation is the difficulty of conducting real-world experiments over a variety of FL client devices and different platforms, with different datasets and data distribution, all while assessing various dimensions of algorithm performance, such as inference accuracy, energy consumption, and time to convergence, to name a few. In this paper, we present \colext, a real-world testbed for FL research. \colext is designed to streamline experimentation with custom FL algorithms in a rich testbed configuration space, with a large number of heterogeneous edge devices, ranging from single-board computers to smartphones, and provides real-time collection and visualization of a variety of metrics through automatic instrumentation. According to our evaluation,  porting FL algorithms to \colext requires minimal involvement from the developer, and the instrumentation introduces minimal resource usage overhead. Furthermore, through an initial investigation involving popular FL algorithms running on \colext, we reveal previously unknown trade-offs, inefficiencies, and programming bugs.
\end{abstract}


\section{Introduction}
\label{sec:introduction}

Data is a most precious resource, and the one that, with the growth of privacy awareness, users tend to be less inclined to share with third parties. Centralized AI has made tremendous advances in the last couple of decades. Yet, virtually all of the publicly available data, such as the content of the World Wide Web, are already being used for large foundational models. The next breakthrough in AI will necessarily have to rely on high-quality private data residing on end-user devices.

Harnessing individuals' data while maintaining privacy is a challenging feat, and various research approaches have been proposed to tackle this issue~\cite{li2023survey,kairouz2021advances}. Federated learning (FL) allows distributed collaborative training of machine learning (ML) models over a group of participating devices, with a centralized server used merely for training orchestration, essentially aggregating clients’ model updates and sharing the newly-created model within the group~\cite{kairouz2021advances}. Its conceptual simplicity makes FL the most popular solution for privacy-preserving distributed AI, especially when deep learning (DL) models are involved.

The original FL algorithm, FedAvg~\cite{mcmahan2017fedavg} was shown to underperform when different assumptions, such as the uniformity of data distribution or computational capabilities over clients, are lifted~\cite{caldas2019leaf,hsieh2020quagmire,yang2021flash,wu2022motley,Abdelmoniem.IoTJ23,baumgart2024federated,zhang2024flhetbench,hsu2019fedavgm}.
A large body of theoretical work has followed, and various attempts to alleviate the issue have been made~\cite{li2020fedprox,li2021moon,lin2020feddf,karimireddy2020scaffold,wang2020tackling,fourati2023filfl}.
Nevertheless, these scientific contributions very rarely trickle down to practice, primarily because they do not provide readily usable implementations and fail to convince that the claimed improvements indeed translate to real-world deployments. Instead, most proposals remain at the level of a simulation running on a single server, and the real-world behavior in terms of the algorithm running time, computational/memory/energy demand, and performance under various real-world constraints, such as with data/network/device heterogeneity, remains unknown.

\begin{figure}[tp!]
    \centering
    \includegraphics[width=\linewidth]{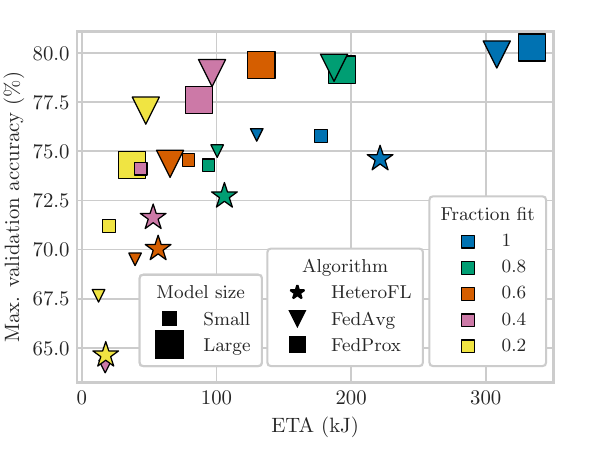}
    \caption{Max. validation accuracy and energy to accuracy (ETA) for three FL algorithms on the CIFAR-10 dataset.
    In FedAvg and FedProx, all clients use either a \textit{Small} or a \textit{Large} model, while in HeteroFL, clients use one of the two depending on the computational power. The ETA axis values cannot be (reliably) assessed without real-world experimentation provided by \colext.}
    \label{fig:eta_vs_max_acc}
\end{figure}

In this paper, we aim to increase the credibility of FL research by providing \colext~-- a solution for reproducible FL experimentation over real-world devices while also enabling a gamut of relevant performance metrics to be collected. Our solution is tailored to support virtually any existing and future algorithm constructed on top of the currently most popular FL framework, Flower~\cite{beutel2022flower}, and enables a range of scenario-defining parameters, such as client heterogeneity, metric collection configurations, and others, to be set. We implement \colext in a federation of 28 single-board computers (SBCs) and 20 Android smartphones and demonstrate its utility for objectively assessing FL algorithms over different dimensions.

As an illustrative example, in \autoref{fig:eta_vs_max_acc}, we depict the result of \colext experimentation over three popular FL algorithms (FedAvg~\cite{mcmahan2017fedavg}, FedProx~\cite{li2020fedprox}, and HeteroFL~\cite{diao2021heterofl}), with two different model sizes (small -- with 35k and large -- with 380k parameters), and a different ratio of clients participating in every training round. Traditional means of simulation-based assessment would ``see’’ only the $y$-axis in the figure, essentially comparing algorithms based on the highest accuracy achieved, and would identify FedProx over a 380k parameter model with full client participation as the most promising solution. \colext, on the other hand, uncovers other metrics that may be relevant, such as the memory, CPU, and energy usage, as well as the training duration, and juxtaposes them with the achieved accuracy. In~\autoref{fig:eta_vs_max_acc}, for instance, \colext reveals that the amount of energy needed for reaching a particular level of accuracy (i.e., energy-to-accuracy, ETA) differs drastically among points that achieve very similar accuracy. Thus, the previously identified FedProx configuration reaches the top accuracy while consuming almost 350 kJ (kilo Joules). At the same time, involving only 40\% of the clients in each training round, the FedAvg algorithm incurs a 4 p.p. (percentage points) decrease in accuracy while consuming less than a third of the energy (i.e., 100 kJ).

The above is just one of the examples of how \colext can uncover nuances related to the performance of FL algorithms in real-world environments and can do this with minimal involvement of the algorithm developer. More broadly, our work brings the following contributions to FL research:

\noindent $\bullet$ \textbf{We design and implement a framework for experimentation with FL algorithms that readily supports a wide range of existing and future FL solutions.} Experimenters need to make minor changes (three lines of code) to run their Flower-ready algorithms on \colext.\\
\noindent $\bullet$ \textbf{We instrument algorithms to collect a range of metrics.} We implement low-level acquisition of CPU/GPU utilization, memory consumption, training time, and network usage. We also expose real-world energy measurements through power meters deployed in our setup.\\
\noindent $\bullet$ \textbf{We develop an expressive interface for defining the experimentation scenario.} An experimenter can decide on the level of heterogeneity among devices, change the experiment's training parameters, and configure metric collection settings.\\
\noindent $\bullet$ \textbf{We deploy \colext in a heterogeneous device setting and perform thorough experimentation with a number of popular FL algorithms.} Our \colext testbed includes both SBCs and Android devices. Through microbenchmarks, we confirm that \colext can support a range of algorithms with negligible impact on the device's resources, while use-case experimentation demonstrates how \colext can be harnessed to uncover implementation inefficiencies and trade-off issues related to the real-world use of FL.

While this paper unveils only a few important and interesting revelations (such as the one described in~\autoref{fig:eta_vs_max_acc}), we believe that \colext will help practitioners navigate the trade-offs that different FL algorithms avail and will also help researchers identify the most promising directions for future development of FL algorithms. To facilitate this, \colext is available to interested researchers and the code is open source. Information on how to access the testbed and the code can be found at \url{https://github.com/sands-lab/colext}.

\section{Background and Obstacles to Realistic FL Experimentation}
\label{sec:background}

\subsection{FL Primer and Algorithm Variations}

FL is arguably the most promising means of training AI models in a distributed manner so that the data of individual training participants (clients) remain private. In its simplest form, FL operates in rounds  $t = 0, 1, \ldots, T-1$, and lets each client $k$ from a set of clients $\mathcal{K} = \{1, 2, \ldots, K\}$ independently train a model $\mathbf{w}$ over its local dataset $\mathcal{D}_k$ using stochastic gradient descent (SGD) or a variation thereof. After $E$ local training epochs, the updated versions of the model $\mathbf{w}_k$ are sent to the server, which then aggregates them in a new version of the global model, usually weighting the contribution of each client according to the number of data samples $n_k$ in $\mathcal{D}_k$ : $\mathbf{w}^{(t+1)} = \frac{1}{\sum_{k \in \mathcal{K}^{(t)}} n_k} \sum_{k \in \mathcal{K}^{(t)}} n_k \mathbf{w}_k^{(t, E)}. $ After $T$ rounds, the global model $\mathbf{w}^{(T)}$ is considered trained.

The above algorithm, which is essentially a form of distributed SGD with local steps, is termed FedAvg and represents the de-facto baseline from which numerous other FL algorithms have been developed and have been pitted against. For instance, FedProx~\cite{li2020fedprox} extends the FedAvg local training loss function $F_k(\mathbf{w})$ with a proximal term $\frac{\mu}{2} \|\mathbf{w} - \mathbf{w}^{(t)}\|^2$, where $\mu$ is a non-negative constant, that penalizes large deviations of the local model from the global model. i.e. the loss function becomes: $F_k^{\text{prox}}(\mathbf{w}) = F_k(\mathbf{w}) + \frac{\mu}{2} \|\mathbf{w} - \mathbf{w}^{(t)}\|^2$. Other solutions may introduce other modifications and innovations, for instance, by changing the way in which local weight updates are aggregated on the server~\cite{lin2020feddf}, enabling knowledge distillation among the global and the local models~\cite{aljahdali2024flashback}, or, as is the case with HeteroFL (results of which are depicted in~\autoref{fig:eta_vs_max_acc}) by allowing aggregation of models of different sizes. Notably, even without any modifications, FedAvg already enables significant customization, as several hyperparameters, such as the number of local epochs, the number of clients per round, the deadline for receiving an update from a client, and the number of clients that have to report to a server for a round to be considered successful, all can be tuned and have been shown to influence the performance of the algorithm~\cite{Abdelmoniem.IoTJ23,zhang2024flhetbench}.

\subsection{Experimentation Challenges}

\subsubsection{Impact of Heterogeneity on FL}

In a centralized setting, SGD is performed on the same computing device, and the data comes from the same distribution in each iteration of the algorithm. When SGD is, through FL, deployed over multiple clients, the convergence of the resulting distributed algorithm may be affected by the heterogeneity of a real-world setting. For practical purposes, critical heterogeneities affecting FL include data, hardware, and platform heterogeneities.

\noindent \textbf{Data heterogeneity.} The nature of data distribution among clients plays a pivotal role in FL. In Independent and Identically Distributed (IID) scenarios, each client's data conforms to a uniform distribution, simplifying model aggregation across devices. Conversely, non-IID data presents a more challenging landscape where client dataset distributions vary significantly. This diversity demands nuanced strategies to adaptively reconcile differences between local and global models while preserving data privacy and achieving robust model performance.

Data heterogeneity is the most broadly examined property of realistic FL, and a plethora of algorithms, including previously described FedProx, have been proposed to address the issue~\cite{li2020fedprox,li2021moon,lin2020feddf,karimireddy2020scaffold,wang2020tackling,fourati2023filfl,aljahdali2024flashback,reddi2021adaptive,haddadpour2019convergence,zhang2021fedpd,hsu2019fedavgm}.
At least a part of the reason for intensive research in this direction can be explained by the ease at which one can experiment with FL over non-IID data -- the actual clients may remain simulated, while datasets assigned to these clients can be made artificially non-IID.

\noindent \textbf{Hardware heterogeneity.} ``Stragglers,'' clients who, usually due to poor computing capabilities, take prohibitively long to complete a round of local training, present a major issue in practical FL~\cite{bonawitz2019tff}. Under the presence of stragglers, FL becomes highly inefficient because either other clients must wait for the stragglers, or the stragglers' local updates must be discarded for the learning to advance~\cite{abdelmoniem2023refl}. Alternatives, such as asynchronous FL, have been proposed~\cite{ortega2023fedbuff}, but a common approach in practice is to select clients with uniform hardware specifications~\cite{lai2021oort}. Unfortunately, such an approach severely limits the applicability of FL and may introduce bias in the resulting models, as clients of certain characteristics (and, consequently, data properties) do not feature in the training.

Solutions for learning over clients of heterogeneous capabilities have been proposed~\cite{diao2021heterofl}, yet independently evaluating such solutions remains challenging, as it would require a testbed of sufficiently heterogeneous clients. Furthermore, with heterogeneous clients come heterogeneous processing speeds, power consumption, memory usage, and other metrics, which suddenly expand the dimensionality in which the optimal FL solution should be sought. As seen in~\autoref{fig:eta_vs_max_acc}, introducing just one dimension (energy to accuracy) may alter the way we assess the optimality of an FL algorithm.

\noindent \textbf{Platform heterogeneity.} FL is promoted as a solution for edge AI. Yet, ``edge'' encompasses a wide range of devices, from embedded devices and single board computers (SBCs) common in the Internet of Things (IoT) deployments, to smartwatches, smartphones, and beyond. Nevertheless, FL algorithms are rarely tested in actual deployments, and even more rarely are evaluated on multiple platforms. Mobile devices, in particular, tend to be highly underrepresented when it comes to evaluating FL solutions. FL over different platforms is challenging to implement, and, to the best of our knowledge, only one framework -- Flower~\cite{beutel2022flower} -- enables distributed training over both Linux and Android platforms.\footnote{Flower does not allow a mix of different platforms in the same FL setting.}

\subsubsection{Experiment Orchestration and Testbed Implementation}

\noindent \textbf{Defining, scoping, and monitoring experiments.} Thorough examination of FL algorithms should encompass experimentation over different datasets and different data distributions, with a varying number of clients involved and reporting, to name just a few experiment parameters that should be considered. Without an easy-to-use support for such experimentation, researchers either limit the richness of the experimental scenarios or develop their own infrastructure for such experimentation, which is both time consuming and error prone.

Furthermore, a realistic view of algorithm performance necessitates its realistic deployment. Constructing a full-fledged hardware testbed requires significant resources, both in terms of time and money. A high-end mobile device can cost about \$1,000, an SBC can cost up to \$500, while a high-frequency power meter costs about \$1,000. Equipping an FL testbed with a few dozen devices can be prohibitively expensive for many smaller research groups to do. Moreover, hardware failures (especially networking connectivity) and software updates require active DevOps effort to maintain testbed usability.

\noindent \textbf{Performance metric collection and analysis.} In Section~\ref{sec:introduction}, we have demonstrated the need for assessing FL algorithms along different dimensions. Metrics, such as CPU/GPU utilization, energy and power consumption, memory consumption, data transfer sizes, and others, can paint a different picture of an algorithm's performance compared to merely inspecting the inference accuracy on a test set. However, capturing these metrics without requiring changes in the algorithm code and without affecting the execution of the algorithm is challenging. Moreover, the data should be collected throughout the experiment, reliably transferred and stored, and presented to the experimenter in an appropriate manner.

\section{Related Work}
\label{sec:related_work}

Realistic experimentation is at the core of computer science research. With the rise of distributed and networked computing, a need for more elaborate testbeds has appeared. Consequently, testbeds tailored to allow multiple researchers to conduct relatively diverse experiments have appeared. Planetlab, for example, was a global testbed for computer networking services research that spawned over more than 1,000 distributed nodes at its peak~\cite{peterson2003planetlab}. Emulab, on the other hand, allows experimentation with various networking topologies that are emulated over a cluster of networking devices~\cite{white2002netbed}. Building upon the Emulab software, the CMU wireless emulator enables remote emulation of wireless propagation conditions~\cite{judd2005using}. Also on the wireless front, the ORBIT testbed allowed indoor and outdoor evaluation of wireless protocols~\cite{raychaudhuri2005overview}, while massive MIMO testbeds, such as~\cite{sakhnini2022near}, enable experimentation with future 5G and 6G wireless transmission protocols.

The above testbeds have had a significant impact on networking research. Yet, besides the networking aspect, practical FL encompasses mobile systems and ML aspects. Testbeds, such as CityLab~\cite{struye2018citylab}, facilitate IoT systems research, however, do not cover mobile, especially smartphone-based, computation, and do not readily support distributed ML applications. ML testbeds, on the other hand, focus on cloud computing (e.g., CloudLab~\cite{duplyakin2019cloudlab}) and do not support FL over edge devices.

FL research is supported by several open-source frameworks that have emerged in recent years~\cite{beutel2022flower,foley2022openfl,roth2022flare,xie2023federatedscope,he2020fedml,yang2021fate,ziller2021pysift,daga2023flame,galtier2019substra,kourtellis2020flaas,chen2023fs-real,tff,ludwig2020ibm-fl}. These frameworks typically provide APIs for users to express DL model architectures, data loading, and model training algorithms. Often, the frameworks separate the client-side training logic from the server-side aggregation logic. In several cases, such as with Flower~\cite{beutel2022flower}, these frameworks include a simulation backend that allows an FL system to be run in a simulated environment without any substantial code changes.

Simulating FL is essential to expedite the prototyping of FL algorithms, system designs and evaluations thereof. Generally speaking, simulation can facilitate studying FL systems in different scenarios where the user can control the number of clients, the conditions in which they operate, the algorithms they execute, and other factors. To aid in simplifying large-scale simulation, several toolkits have been proposed to support simulating FL workloads~\cite{zhang2023fedhc,lai2022fedscale,garcia2022flute,yang2021flash,li2021flsim,zeng2022fedlab,ro2021fedjax,mugunthan2020privacyfl,zhuang2022easyfl}.
To supplement FL simulation with realistic characteristics of heterogeneous devices, Protea~\cite{zhao2022protea} proposes to profile devices to obtain information regarding resource consumption and computation time. Similarly, FedScale~\cite{lai2022fedscale} builds a simulation infrastructure on top of realistic client device behavior traces in order to account for system-level heterogeneities that might affect FL. Nevertheless, certain aspects of FL performance, for instance, individual devices' energy consumption, cannot be reliably assessed through trace-based simulation due to the use of simplified models~\cite{ahmad30survey_energy_smartphones,guo2021survey_energy_emb_systems}. 

Critical for FL experimentation is the need to take real-world heterogeneities into account. Several studies have analyzed heterogeneity at the level of data~\cite{nilsson.DIDL18,reddi2021adaptive,selialia2022bias,hsieh2020quagmire},
system~\cite{baumgart2024federated,zhang2024flhetbench}, and client availability~\cite{Abdelmoniem.IoTJ23,yang2021flash}, and have shown that these factors significantly impact the performance of FL. Consequently, several benchmarks that include comprehensive data partitioning strategies to cover the typical non-IID data cases have been introduced~\cite{li2022silos,terrail2023flamby,zhang2023fedlegal,hu2022oarf,song2022flair,wu2022motley,hsu2020federated,caldas2019leaf}.

Nevertheless, to this date, most FL research results have been obtained through simulation, and only relatively few studies (e.g.,~\cite{sun2022fedmsa,shin2022fedbalancer,ouyang2021clusterfl,mills2020ce-fedavg,woisetschläger2023fledge}) make use of actual experimental testbeds. Wong et al. perform a thorough study of FedAvg on a real-world edge computing testbed~\cite{wong2023empirical}. Their study demonstrates the utility of experimentation over heterogeneous hardware (the authors use Raspberry Pi 3/4, Jetson Nano and Jetson TX2), with various experiment settings (e.g., different data distributions), while different metrics (such as CPU utilization) are collected. Grounded in these findings, but not limited to FedAvg, \colext enables real-world experimentation on a wide span of hardware platforms, including 28 SBCs across 6 hardware types and 20 Android mobile phones across 8 models and 5 vendors, all while a wide range of metrics, from CPU utilization, to the amount of transferred data, to high-frequency energy measurements are collected.

\section{\colext: Federated Learning Testbed}
\label{sec:architecture}

\begin{figure}[tp]
    \centering
    \includegraphics[width=\linewidth]{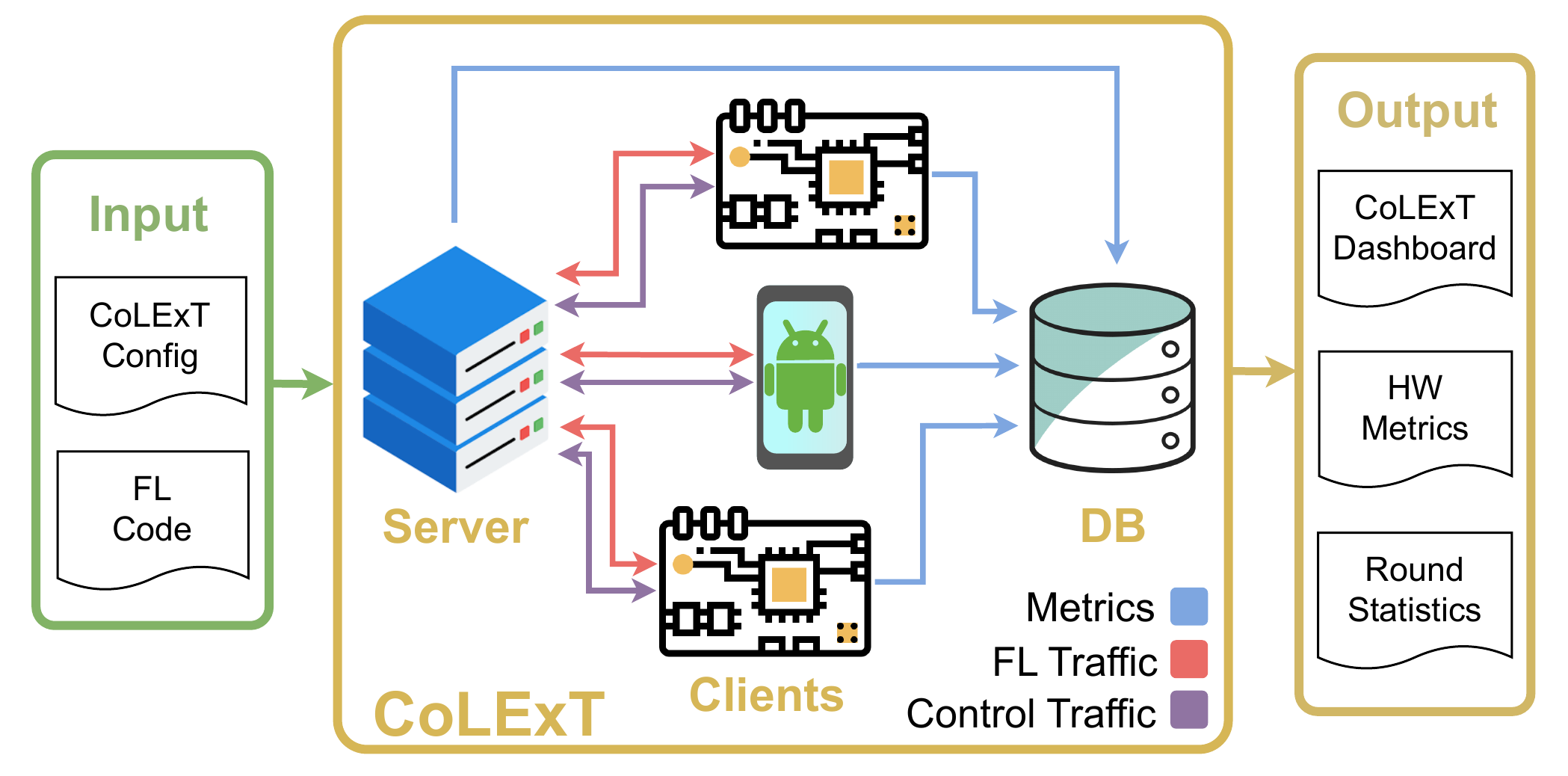}
    \caption{\colext\ workflow.}
    \label{fig:colext_diagram}
\end{figure}

In this paper, we develop \colext, a testbed for comprehensive experimentation with FL over real-world devices. \colext is designed as a solution for seamless deployment of arbitrary FL algorithms and supports a range of experimental scenarios. Characteristically, the testbed supports highly heterogeneous deployments and a comprehensive set of performance metrics.

The overview of the testbed is shown in \autoref{fig:colext_diagram}. The \textbf{\colext server} acts as a central entity for both experiment coordination as well as model aggregation within an FL algorithm. The experiment configuration, \textbf{\colext Config}, is provided by the experimenter. This configuration includes a reference to the code of the FL algorithm under test, \textbf{FL Code}, and the selection of \textbf{\colext clients} that will participate in the experiment. Clients can be selected from a range of devices, which, in the current implementation of \colext, include ARM and x86 SBCs (some having CUDA-capable GPUs) and Android smartphones.

The \textbf{\colext server} instantiates the experiment by packaging experimenter-provided FL code into deployable units and deploying these to the appropriate devices: for SBCs, \colext packages the code into containers and deploys them with Kubernetes; for Android smartphones, the code is integrated within the \colext Android app code, packed into an APK, and deployed with Android Debug Bridge (ADB). The \textbf{\colext server} also executes the FL server code provided by the experimenter.
Separately, the experiment devices collaborate to collect relevant metrics. These are collected partly on clients and partly on the server and are transferred to the \textbf{\colext database} for further inspection.

Once the experiment is completed, the researcher will find a rich range of hardware and system metrics and statistics to gain insights into the performance of their FL algorithm, compare different FL algorithms based on the metrics, and even identify the performance difference across various devices. To facilitate the analysis, \colext also provides a Grafana-based \textbf{\colext dashboard }where the metrics collected during an experiment can be visualized in real-time.

\subsection{Using \colext in a Nutshell} \label{sec:using_colext}

Interaction with \colext is designed to be succinct and in line with the workflow employed when using currently popular FL simulation environments~\cite{beutel2022flower,li2021flsim,roth2022flare,garcia2022flute}. In a nutshell, the experimenter has to:
\begin{enumerate}
    \item Access the \colext server running Python, and in a local Python environment, install our \texttt{colext} package.
    \item {In the FL code, import the above library and wrap the FL client and strategy code with \colext\ decorators. If used outside of the testbed, these decorators do not modify the program behavior and thus can safely be included in the code in general.
    An example of decorated client and strategy would be:
\begin{listing}[H]
\begin{minted}
[
frame=lines,
framesep=2mm,
baselinestretch=1.2,
fontsize=\scriptsize,
]
{Python}
from colext import MonitorFlwrClient, MonitorFlwrStrategy

@MonitorFlwrClient
class FlowerClient(fl.client.NumPyClient):
[...]
@MonitorFlwrStrategy
class FlowerStrategy(flwr.server.strategy.Strategy):
[...]
\end{minted}
\end{listing}
}
\item Declare the pip \texttt{requirements.txt} file with the dependencies.
\item {Declare the \texttt{\lowercase{\colext{}\_config.yaml}} experiment configuration file, which specifies the client and server entry points and the testbed devices on which the experiment will run. An example of such a file would be:

\begin{listing}[!ht]
\begin{minted}
[
frame=lines,
framesep=2mm,
baselinestretch=1.2,
fontsize=\scriptsize,
]
{YAML}
code:
  client:
    entrypoint: "client.py"
    args:
        - "--server_addr=${COLEXT_SERVER_ADDRESS}"
        - "--client_id=${COLEXT_CLIENT_ID}"
  server:
    entrypoint: "server.py"
    args: "--n_clients=${COLEXT_N_CLIENTS} --n_rounds=3"
devices:
  - { dev_type: LattePandaDelta3, count: 4 }
  - { dev_type: OrangePi5B,  count: 2 }
  - { dev_type: JetsonOrinNano, count: 4 }
monitoring:
  scrapping_interval: 0.3 # in seconds
  push_to_db_interval: 10 # in seconds
\end{minted}
\end{listing}}

\item {Deploy the experiment using the \texttt{colext\_launch\_job} command, after which the experiment performance metrics will be available for real-time monitoring on the \colext dashboard. Optionally, after the experiment is completed, the experimenter can call \texttt{colext\_get\_metrics} to retrieve the collected data in the form of CSV files from the \colext database.
In the terminal, it would look like:

\begin{listing}[!ht]
\begin{minted}
[
frame=lines,
framesep=2mm,
baselinestretch=1.2,
fontsize=\scriptsize,
]
{Bash}
$ colext_launch_job --config <path-to-config>
# Prints a job-id and a Grafana dashboard link

# After the job finishes, retrieve metrics for job-id
$ colext_get_metrics --job_id <job-id>
\end{minted}
\end{listing}
}

\end{enumerate}

\section{\colext Implementation}
\label{sec:implementation}

\subsection{Underlying FL Framework}
\colext is designed to provide a realistic picture of FL algorithm performance in a real-world deployment, yet, at the same time, it aims to minimize the effort one needs to put into the experimentation. Therefore, rather than developing a custom solution for FL, we base \colext on an existing framework, offering researchers a convenient way to deploy their existing code with minimal modification.

From the available frameworks, we opted for Flower~\cite{beutel2022flower}. This framework offers a simple interface that facilitates FL research and is currently the most popular FL framework, with a large community of developers actively using and developing the framework.\footnote{Flower GitHub page has been ``starred'' 4.4K times, while the TesorFlow Federated page has accumulated 2.3K stars. In addition, the Flower website states that the framework is associated with ``The world's largest Federated Learning conference" -- Flower AI Summit 2024.} Moreover, the Flower community led a substantial effort towards reproducing FL research, which has generated a pool of high quality baselines coded against the Flower API. From the technical side, Flower supports synchronous FL, relying on gRPC\cite{grpc} protocol, and ensures that on-client training, on-server aggregation, and communication between the FL clients and the server are executed.

While \colext relies on Flower, it is by no means locked into using this framework. Indeed, Flower could be substituted with any other framework that supports on-device execution of FL, as long as, besides the FL training control commands, the framework exposes API hooks for indicating the beginning/end of an FL round at the server and the beginning/end of the local training on the client.

\subsection{\colext Client and Server}
\label{sec:implementation:client}

The majority of FL research has been developed in Python and evaluated on Linux-based x86 machines due to the ease of use of this platform. Real-world FL deployments, on the other hand, are expected to include other platforms, such as the Tegra architecture of NVIDIA Jetsons, or even different operating systems, such as Android. With \colext, we aim to support code execution on different devices with minimal effort from the experimenter's side.

Android and Linux-based clients are handled differently in Flower. Consequently, in \colext we devise a separate Linux FL client and an Android FL client.

\textbf{Linux client} leverages containers cross-compiled using Docker Buildx to support multiple architectures, including AMD64 and ARM64. These containers are then stored on our private container registry using Harbor\cite{goharbor}. \colext expects that experimenters have their FL code written in Python and will use the supplied pip \texttt{requirements.txt} file to install the dependencies within the container.

Due to the unavailability of specific Python packages on the Python Package Index (PyPI) for some of the SBC architectures used in our testbed, we manually pre-compiled multiple versions of certain Python packages (such as PyTorch with GPU support for ARM) and enabled their inclusion during the container packaging process.

In the client code, \colext expects that \colext decorators have been applied to automatically collect performance metrics from the FL code.
Finally, while the common situation of an experimenter providing the Python code with the pip \texttt{requirements.txt} file is handled through automatic container packaging (the configuration specification is described in \autoref{sec:using_colext}), other setups, such as those involving custom dependency compilation, are possible with manual container building.

\textbf{Android client} is based on the \colext Android app, within which an experimenter copies their Java/Kotlin code defining the client behavior. The provided app exposes the TensorFlow Lite support for on-device DL training, handles communication with the server, and collects performance metrics.

\textbf{FL server} is provided as a Python script, irrespective of whether the clients are SBCs or smartphones, and is expected to run on a Linux x86 machine. Similar to the Linux clients, the FL server code is containerized with all its dependencies installed. \colext expects the server code to utilize the server-side \colext decorator to collect performance metrics, e.g., round timings. Additionally, the container is configured to have access to the GPU on the host machine in case the server code can benefit from the accelerator.

\subsection{Datasets and Data Partitioning}

Regardless of client type, clients must obtain their dataset at the start of each experiment. To avoid repeated downloads, commonly used datasets in FL research, such as CIFAR~\cite{cifar10} and MNIST~\cite{mnist}, are cached on all devices. Users can include additional datasets to be cached if needed.

Note that since the entire dataset is installed on each device, clients still need to determine which data points comprise their specific subset. To stay general, \colext is oblivious to specific data partitioning schemes, and we assume the responsibility of devising data partitions lies with the user.\footnote{For instance, one option is to assign dataset subsets to clients by generating a CSV file for every client, listing the data point indices to be used, as shown in \url{https://github.com/sands-lab/flower_dcml_algorithms}.}

\subsection{Collecting Performance Metrics}

In \colext, we aim to seamlessly capture experiment performance metrics with minimal modification to the FL research code and also support intuitive and informative visualization of the captured metrics.

The performance of FL has traditionally been evaluated along a single dimension -- inference accuracy of the resulting model. \colext naturally supports the collection of this metric by logging the returned accuracy values from Flower's server and client evaluation functions. Nevertheless, as shown in our introductory example in \autoref{fig:eta_vs_max_acc}, metrics, such as the energy consumption per device, execution time, and others, are necessary for a holistic evaluation of FL algorithms. In \colext, such metrics are collected by a background scraper that periodically records them, timestamps them with the local time, and aggregates the data in batches before sending them to \colext database. The measurements are unrelated to FL rounds. However, as the devices in the testbed are synchronized using NTP, \colext can associate the performance data with rounds by grouping the data according to the round start/finish time.

The way the metrics are collected differs between Linux and Android clients. \textbf{In Linux}, standard hardware metrics such as CPU, memory, and network utilization are obtained using the \texttt{psutils} Python package from a separate background process. Power consumption, however, is tracked differently for different devices. Thus, for NVIDIA Jetsons, we obtain the power consumption of the entire board using \texttt{jetson-stats} monitoring tool. For LattePanda devices, on the other hand, such a tool is not available, and we capture the CPU power consumption using Intel ``Running Average Power Limit" (RAPL) through the \texttt{pyRAPL} Python package. Finally, for OrangePi devices, no suitable software solution exists; thus, we resort to physical power consumption measurement using the Monsoon power meter~\cite{powermonitor}.

Regarding GPUs, even though all SBCs have one, we were only able to train ML models on NVIDIA Jetsons' GPUs. The other GPUs, the Intel HD Graphics GPU on the LattePanda and the Mali GPU on the Orange Pi, lack support from major ML frameworks, as these GPUs primarily focus on graphics processing, thus have limited memory and lack support for precision formats required for machine learning. NVIDIA Jetson integrated GPUs, on the other hand, support CUDA and, for the purpose of deep learning, can be treated as discrete GPUs. Metrics for NVIDIA Jetson GPUs were collected using the \texttt{jetson-stats} package.

\textbf{In Android}, accessing performance metrics is more challenging than in Linux. While in the past, parsing \texttt{/proc/stat} allowed one to retrieve device utilization metrics, in newer versions of Android (as of Android 12), this is not available anymore. We therefore develop a set of techniques for accessing various metrics on Android devices. First, we assign special privileges to our app in Android OS, where we register the app as the device owner. Then, we use the official memory utilization API to obtain the memory usage and \texttt{/sys/devices/system/cpu} scraping for CPU utilization statistics. Obtaining GPU usage statistics is highly challenging, as there is no official API for GPU statistics on Android, nor do the usage statistics files necessarily exist on the file system. Our investigation with multiple phone makers, and models (Google Pixel 7, Xiaomi 12, Samsung Galaxy S21 FE, M54, XCover 6 Pro, ROG Phone 6, OnePlus Nord 2T 5G, and Xiaomi Poco X5 Pro) finds that only Samsung devices reliably expose the GPU utilization files, and only in case the devices are rooted. We, thus, collect GPU statics for Samsung devices within the \colext testbed. Finally, when it comes to power consumption, the official Android \texttt{BatteryManager} API allows us to collect power consumption attributed to different apps and the system as a whole, all from our app that was previously given device owner privileges.

\subsection{Experiment Orchestration}

\colext automates the deployment and running of FL experiments on real-world clients. Since the underlying FL framework handles Android and Linux clients separately (see also Section~\ref{sec:implementation:client}), the experiment orchestration varies between SBCs and smartphones.

\textbf{For SBCs}, we create a Kubernetes cluster containing Linux clients and the server. We use the \texttt{microk8s} orchestration tool as it is geared towards edge devices and conveniently available as a \texttt{snap} package, which isolates the Kubernetes installation from the host filesystem while natively running the entire Kubernetes stack without the need for containers. Container deployment in Kubernetes requires configuring pods, the deployment unit in Kubernetes. \colext\ prepares FL client and server pod configurations using \texttt{Jinja2} template files that can be configured with the required device type for the client, entry point, mounted directories for dataset caching, and added \colext\-related environment variables (listed in \autoref{tab:env_vars}), including a client identifier and device type.

Exposing and utilizing GPU computation through a Kubernetes container is done through \texttt{microk8s}, yet the support is limited to discrete GPUs and is incompatible with integrated GPUs of NVIDIA Jetsons. To overcome this, we configured the underlying container runtime on those devices to use the NVIDIA container runtime as the default runtime, which exposes the GPU to any containers running on the device.

Finally, we also had issues with the default \texttt{microk8s} Container Network Interface (CNI), Calico, because the \texttt{ipset} kernel module is missing from Jetsons. To avoid this issue, we switched to another CNI, Flannel~\cite{microk8s_issue}.

\begin{table} [t]
    \centering
    \caption{\colext\ environment variables.}
    \label{tab:env_vars}
    \begin{tabular}{|c|c|}\hline
 Environment Variable&Description\\ \hline
         COLEXT\_SERVER\_ADDRESS& Server address (host:port)\\ \hline
         COLEXT\_N\_CLIENTS& Number of clients\\ \hline
         COLEXT\_CLIENT\_ID& Client ID (0...n\_clients)\\ \hline
         COLEXT\_CLIENT\_DEV\_TYPE& Client device type\\ \hline
    \end{tabular}
\end{table}

\textbf{For smartphones}, a Kubernetes-based solution is not an option. Instead, we build our own deployment system using a combination of Python scripts, Bash scripts, and ADB. ADB allows us to administer Android devices and issue commands for installing and running our applications, while Python scripts, along with Bash commands, provide an interface between \colext server (also written in Python) and deployment scripts. The smartphones are connected to the server in Debug mode, which allows us to transfer files, install the refreshed application (if needed), and run the application for the clients through ADB commands.

\subsection{\colext Dashboard}
\begin{figure}[tp]
    \centering
    \includegraphics[width=\linewidth]{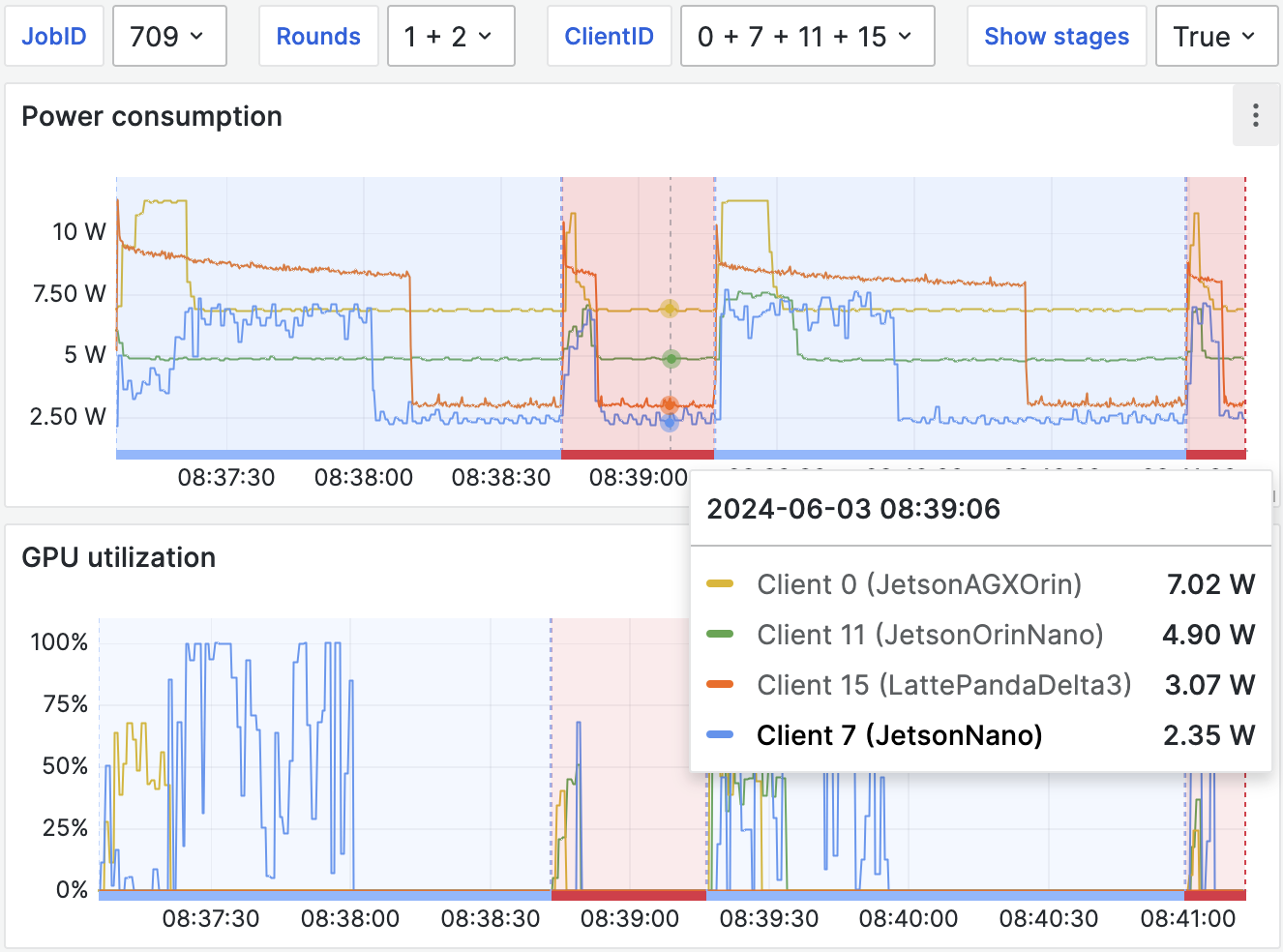}
    \caption{Example capture of \colext\ Dashboard.}
    \label{fig:dashboard_example}
\end{figure}
\colext Dashboard provides a visual depiction of the collected performance metrics. To avoid cluttering the dashboard, an experimenter can limit the metrics to a subset of rounds and clients. A cropped screenshot of the dashboard can be seen in \autoref{fig:dashboard_example}. Hovering over the graph provides information across different clients, while the x-axis denotes time.

The dashboard also allows highlighting periods the FL algorithm spends on different training stages (``Show stages"): training in blue, and evaluation in red. By clearly distinguishing between these two phases, the dashboard helps identify patterns of each phase and reveals the cause of metric spikes (power and GPU utilization) that are present in the example depicted in \autoref{fig:dashboard_example}.

In addition to per-client metrics, the dashboard also contains a section with aggregated metrics over device types to assist with cross-device comparison. Finally, to simplify debugging, \colext can also collect logs and display them directly in the dashboard while the experiment is running.

\section{\colext Testbed Deployment}

\begin{figure}[tp]
    \centering
    \includegraphics[width=\linewidth]{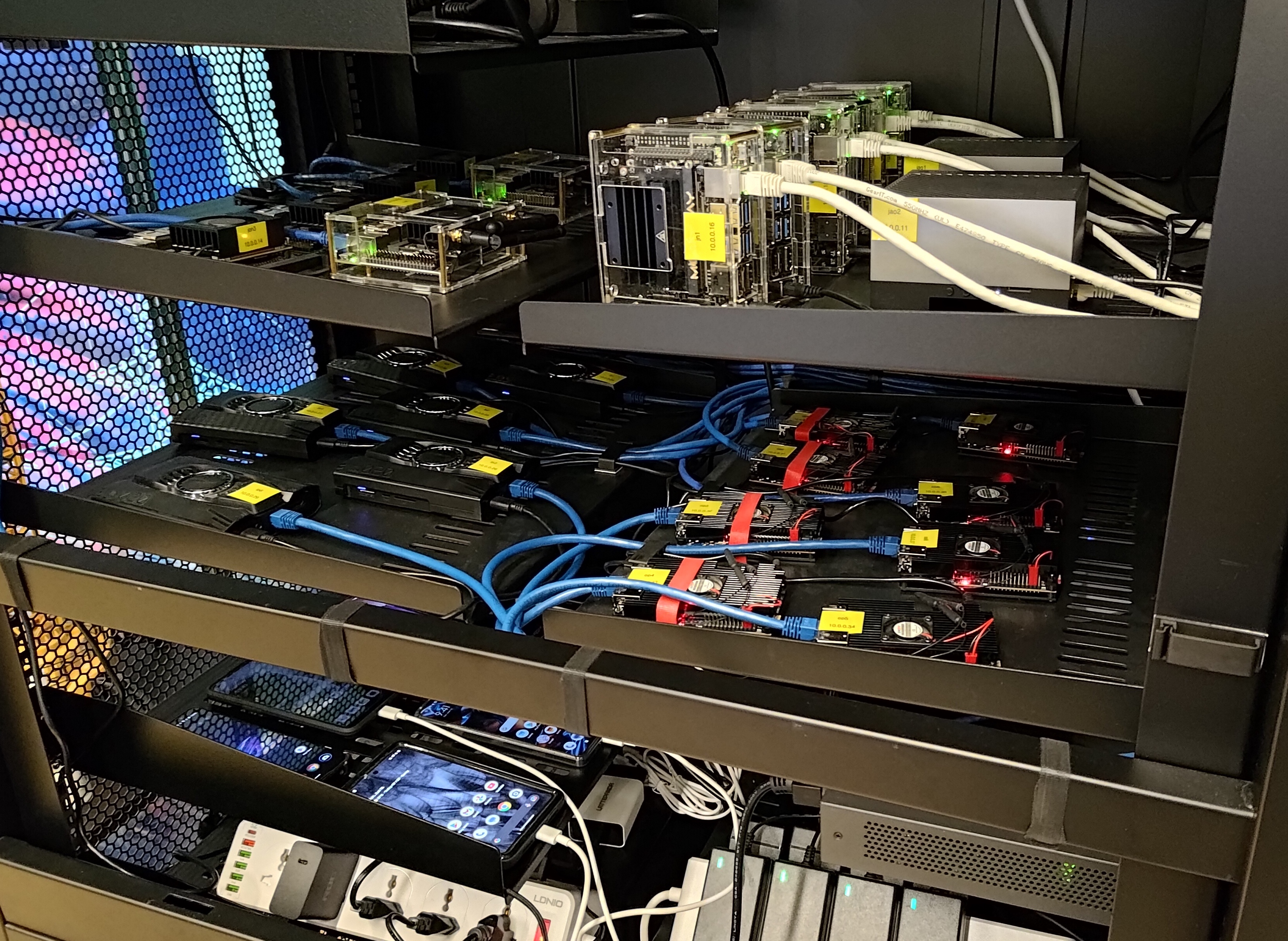}
    \caption{\colext testbed devices.}
    \label{fig:testbed_picture}
\end{figure}

\begin{figure}[t]
     \centering
     \includegraphics[width=\linewidth]{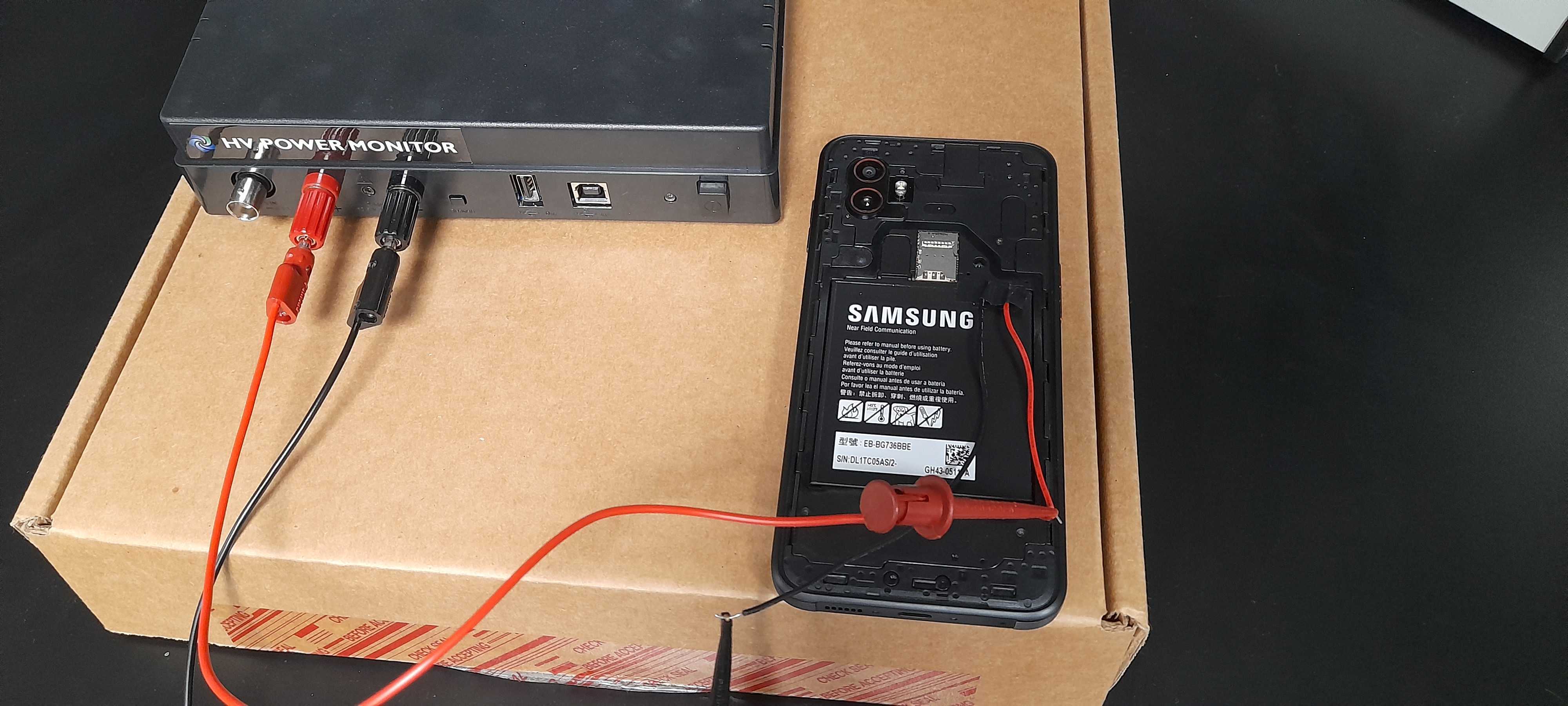}
     \caption{\colext Samsung Galaxy XCover 6 Pro powered by a Monsoon PM.}
     \label{fig:monsoon_picture}
 \end{figure}

We deployed \colext testbed in a dedicated server room at our institution premises. The testbed comprises of heterogeneous edge devices, including SBCs with integrated GPUs, such as the NVIDIA Jetsons, x86 SBCs, such as LattePandas, ARM-based SBCs, OrangePis, and Nvidia Jetsons, and Android devices, selected based on their AI task performance scores from the benchmark AI-Benchmark~\cite{aibenchmark_ranking}, so to cover low-, middle-, and high-end phones. In total, 48 devices, of which 28 are SBCs of 6 model types, and 20 Android smartphones of 8 models (5 vendors), are included in the testbed.  More details, including the quantity of each device type, are present in \autoref{tab:dev_spec}, while a picture of the testbed can be seen in \autoref{fig:testbed_picture}. Furthermore, a server-grade machine equipped with an Intel(R) Xeon(R) Gold 6442Y CPU (48 cores @ 2.6GHz), 256GB of RAM, and an NVIDIA RTX A6000 GPU acts as the \colext server. Finally, the testbed also includes a Monsoon High Voltage Power Monitor (PM), which acts as a power source, while simultaneously monitoring the amount of power supplied to the device.
PM measures current and voltage at a sampling rate of 300 kHz. PM can power SBCs directly through appropriate pins. For Android devices, the battery must be removed and the PM's wires need to be connected to the battery pins.
The Samsung Galaxy XCover series, with its removable batteries, offers the most straightforward implementation of the above, depicted in \autoref{fig:monsoon_picture}.

\begin{table*}[t]
     \centering
     \caption{Testbed device specifications. SBC's AI Performance was retrieved from NVIDIA Jetson Benchmarks~\cite{jetson_benchmarks}. Score values for Android devices were obtained from AI-Benchmark~\cite{aibenchmark_ranking}. Unavailable scores are shown with a dash.}
     \label{tab:dev_spec}
     \begin{tabular}{|l|c|c|c|c|c|}\hline
  \multicolumn{6}{|c|}{Single Board Computer (SBC)} \\\hline
  Device & Qty & CPU (Core@GHz) & Mem (GB@GHz) & AI Perf (TOPS)& OS environment  \\ \hline
Jetson AGX Orin&2&12@2.2&  64@3.2 &  275& Jetpack 5.1.2 + Linux for Tegra 35.4.1  \\ \hline
Jetson Orin Nano&4&6@1.5& 8@2.1 & 40& Jetpack 5.1.2 + Linux for Tegra 35.4.1  \\ \hline
Jetson Xavier NX&2&6@1.9&  8@1.8 &  21& Jetpack 5.1.2 + Linux for Tegra 35.4.1  \\ \hline
 Jetson Nano& 6& 4@1.5& 4@1.6 & 0.472&Jetpack 5.1.2 + Linux for Tegra 32.7.4 \\ \hline
Latte Panda Delta 3&6& 4@2.9& 8@2.9 &--& Ubuntu server 22.04  \\\hline
Orange Pi 5B&8& 8@2.3& 16@2.3 &--& Ubuntu server 22.04\cite{orangePi_OS} \\\hline
  \multicolumn{6}{|c|}{Android Mobile Phone} \\\hline
  Device & Qty & SoC & Mem (GB) & Phone score / SoC score & OS environment  \\ \hline
Asus ROG 6 &2& Snapdragon 8+ Gen 1 & 16 & 1447 / 1000 & Android 13  \\ \hline
Xiaomi 12 &2& Snapdragon 8 Gen 1 & 12  & 1355 / 1046 &Android 12 \\\hline
Google Pixel 7 &5& Google Tensor G2 & 8  & 720 / 525 & Android 13 \\\hline
Samsung XCover 6 Pro &3& Snapdragon 778G & 6  & \hspace{0.8em}--\hspace{0.3em} / 257& Android 13 \\ \hline
Xiaomi Poco X5 Pro & 2& Snapdragon 778G & 8  & \hspace{0.8em}--\hspace{0.3em} / 257&Android 12 \\\hline
Samsung Galaxy S21 FE &2& Exynos 2100 & 8  & 262 / 196 & Android 13 \\ \hline
 OnePlus Nord 2T 5G & 2& Dimensity 1300 & 8  & 240 / 177 &Android 12 \\\hline
 Samsung Galaxy M54 & 2& Exynos 1380 & 8  & \hspace{0.4em}--\hspace{0.3em} / 75&Android 13 \\\hline
\end{tabular}
\end{table*}

The testbed operates over its dedicated network. \colext devices are connected through a switch and a WiFi access point (AP) -- the 28 SBCs, the FL server, and the WiFi AP are connected to a switch using Ethernet cables, while the 20 smartphones connect to the AP via WiFi. The FL server runs a DHCP service, configured with statically assigned IPs,  and is the gateway to the Internet.

The testbed devices are pre-configured with the appropriate operating system and software environment. For SBCs, we opt for Ansible playbooks to automate the configuration of the devices, including the configuration of the NTP server and the installation and configuration of microk8s, which also adds the node to the Kubernetes node pool. For Android, some device configuration is possible through ADB, including the NTP server setup, but other configurations, like the connection to ADB, require manual intervention. Due to hardware differences, the OS and the environment vary across SBCs. The OS also varies for Android devices, depending on the vendor. The OS environment for every device type is described in \autoref{tab:dev_spec}.
The testbed devices are pre-configured with the appropriate operating system and software environment. For SBCs, we opt for Ansible playbooks to automate the configuration of the devices, including the configuration of the NTP server and the installation and configuration of microk8s, which also adds the node to the Kubernetes node pool. For Android, some device configuration is possible through ADB, including the NTP server setup, but other configurations, like the connection to ADB, require manual intervention. Due to hardware differences, the OS and the environment vary across SBCs. The OS also varies for Android devices, depending on the vendor. The OS environment for every device type is described in \autoref{tab:dev_spec}.

\section{Validating \colext}
\label{sec:evaluation}

We validate \colext for its ability to provide the metrics of interest without significant overhead and its ability to support a range of FL algorithms out-of-the-box.

Throughout this section, we use the CIFAR-10 dataset~\cite{cifar10}, which is widely employed in FL research and hence allows us to validate the results we obtain. We also use relatively small models (35k to 380k parameters). Nevertheless, \colext is agnostic to both the models and the datasets the clients use.
Lastly, devices use their default performance configurations, with the exception of NVIDIA Jetsons, whose power mode is set to the highest setting to optimize performance.

\subsection{Quantifying Metric Collection Overhead}
\begin{figure}[tp]
    \centering
    \includegraphics[width=\linewidth]{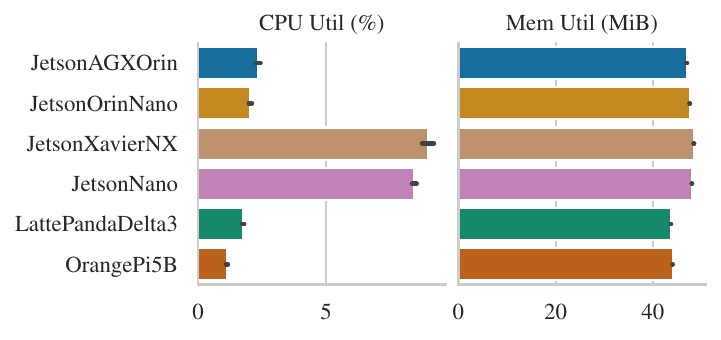}
    \caption{The overhead of performance metric collection on SBCs. The results show average CPU and memory utilization over 1,000 metric collection events, with error bars indicating the 95th percentile. On each platform the CPU utilization remains below 9\%, while the memory utilization remains very low ($<50$MiB).}
    \label{fig:metric_scrapping_overheads}
\end{figure}

Frequent sampling of performance metrics may require significant resources of the host machine, which, in turn, could affect the execution of the FL task. To ensure this is not the case in \colext, we profile the metric scrapper on SBCs. The profiling ran for 5 minutes, scrapping every 0.3 seconds and pushing metrics to the database every 10 seconds. In total, we collected 1,000 sample points. We assess the excess CPU and memory usage caused by such scraping in \autoref{fig:metric_scrapping_overheads}.

For most devices, we see that CPU usage stays below 2.5\%, i.e., remains rather insignificant. The lowest CPU utilization is associated with the OrangePi, which only collects metrics from the \texttt{psutils} package. The second lowest utilization belongs to the LattePanda devices, where, on top of \texttt{psutils}, \colext resorts to \texttt{PyRAPL} to collect power measurements, hence requiring additional CPU utilization. Jetson AGX Orin and Orin Nano experience increased CPU usage, as these devices call \texttt{psutils} and then \texttt{jetson-stats} for power measurements and GPU utilization. Finally, we see a rather unexpected spike in CPU usage for Jetson XavierNX and Jetson Nano, where the CPU utilization is over 3 times higher than for the other Jetson devices, despite using the same data collection method. After profiling the code and benchmarking the CPUs with the \texttt{sysbench} program, we discovered that the CPUs of these devices are approximately twice as slow as the CPUs of other Jetson devices. However, the CPU utilization of XavierNX can be reduced by using a power mode that favors CPU speed by using fewer cores (6@1.4GHz or 2@1.9GHz). Nevertheless, we believe that with less than 10\% CPU overhead, the burden imposed by fine-grain metric collection remains acceptable irrespective of the device type, while the rate of metric collection can be reduced in case lowering the overhead is necessary.

In \autoref{fig:metric_scrapping_overheads}, we also depict the memory usage, which is shown to stay consistently low across devices when metrics are periodically pushed to the DB.  However, if metrics are not periodically pushed, they are stored in memory, causing memory utilization to increase by about 10 KiB for every 1k samples collected, as each sample (of all the metrics) requires an average of 10 bytes. Related to this, the network usage is also determined by how frequently metrics are pushed to the DB. Given the small size of the metrics, only minimal bandwidth is required. Moreover, potential interference of data transmission and the operation of the FL experiment can be entirely avoided by sending the metrics to the DB only after the experiment has finished.

\begin{table}[t!]
    \centering
    \caption{FL algorithms tested on \colext SBCs.}
    \label{tab:ported_algorithms}
    \resizebox{\linewidth}{!}{%
    \begin{tabular}{|l|c|l|}\hline
        Algorithm &  Deployed &  Issues\\\hline
        FedAVGm~\cite{hsu2019fedavgm}  &  \checkmark &  Aarch64 \& TensorFlow on LattePanda\\ \hline
        FedProx~\cite{li2020fedprox} & \checkmark &  -\\ \hline
        Moon~\cite{li2021moon}  & \checkmark&  GPU only\\ \hline
        FedNova~\cite{wang2020tackling} & \checkmark& -\\\hline
        FedPara~\cite{hyeon2021fedpara}& \checkmark&Dataloader segfault LattePanda + OrangePi\\\hline
        HeteroFL~\cite{diao2021heterofl}  & X &  Code decoupling needed\\\hline
        FjORD~\cite{horvath2021fjord}    & X &  Unsupported serialization\\\hline
    \end{tabular}
   }
\end{table}

\subsection{Supporting Different FL Algorithms} \label{porting_baselines}

\subsubsection{Porting FL algorithms on SBCs}

To confirm \colext{}'s ease of use and compatibility with different FL algorithms, we execute a collection of FL algorithms on our \colext testbed. We select these algorithms from open-source implementations developed within the Flower ``Summer of Reproducibility" initiative~\cite{flower_summer}, during which a cash reward was provided for the Flower-based implementations of published FL algorithms. Currently, this collection includes 20 baselines, of which seven (listed in \autoref{tab:ported_algorithms}) were used for our experiments.

These baselines are written so that their folder structure and code organization are uniform, which makes comparison across algorithms straightforward. Nevertheless, due to varying authorship, the coding style, code efficiency, and package dependencies varied noticeably among the algorithms. Thus, we believe that with this set of algorithms, we can comprehensively test the \colext{}'s ability to support different FL algorithms. Note, however, that all of the baselines were initially implemented with a simulation environment in mind, and consequently, certain modifications are necessary to get them working in a real-world client-server deployment. In case the algorithms do not assume shared data between clients and the server, and when the data communicated with the server can be serialized by the client, the required changes are minor. In all cases, the entry script needs to be updated to support the separate start-up of client and server as follows:

\begin{listing}[H]
\begin{minted}
[
frame=lines,
framesep=2mm,
baselinestretch=1.2,
fontsize=\scriptsize,
]
{Python}
# Before: Simulation
history = fl.simulation.start_simulation(
    client_fn=client_fn,
    num_clients=cfg.num_clients,
    config=fl.server.ServerConfig(cfg.num_rounds),
    strategy=strategy,
)

# After: Client - Server
#   is_client, is_server, server_addr, num_rounds, client_id:
#   passed as arguments via the configuration file
if is_client:
    fl.client.start_numpy_client(
        server_address=cfg.server_addr,
        client=client_fn(cfg.client_id),
    )
elif is_server:
    fl.server.start_server(
        server_address="0.0.0.0:8080",
        config=fl.server.ServerConfig(cfg.num_rounds),
        strategy=strategy,
    )
\end{minted}
\end{listing}

As we identify in \autoref{tab:ported_algorithms}, we successfully complete the experiments with five out of seven selected algorithms -- two baselines do not comply with the above requirements and hence cannot be deployed on the testbed without corrections to their codebase. The process was relatively straightforward, and besides the expected augmentation of the code (i.e., 3 lines of code for decorating the FL client and strategy, detailed in Section~\ref{sec:using_colext}, and the above-shown changes to the entry script), we had to take the following additional steps: 1) Flower uses Poetry~\cite{python_poetry} to specify dependencies, thus we had to convert the list of dependencies into the format \colext supports, i.e., the pip \texttt{requirements.txt} file, using the \texttt{poetry export} command; 2) package dependencies for aarch64 sometimes needed adjustments. From our analysis, all baselines were tested on an x86 machine. However, when we try to use the same (identical) dependencies, we sometimes find that minor versions of packages are not available on aarch64. When we encounter these issues, we bump the minor version for one that supports aarch64; 3) we had to replace the use of dependencies explicitly targeting x86 architectures. In Flower baselines, torch and torchvision dependencies are specified by prebuilt wheels targeting x86. We had to comment out these dependencies so that aarch64 wheels could be used for our aarch64 devices.

We now describe the challenges encountered when porting the baselines.  For FedAVGm, we had to degrade TensorFlow from 2.11.1 to 2.11.0 because a dependency required by the original version was not available for aarch64. Additionally, the PyPI TensorFlow wheel for x86 CPUs assumes support for the AVX instruction, which is not available on the LattePandas, so they cannot use those wheels.  Certain parts of the Moon baseline code assume that a GPU is available for the movement of data to/from the GPU. This prevents the experiment from running on CPU-only devices. FedPara is deployable, but we encountered issues with the PyTorch DataLoader module, causing segmentation faults on OrangePis and LattePandas. The exact cause remains unclear, but it could be related to the data partitioning strategy not supporting a reduced number of clients. 
The HeteroFL baseline requires some code decoupling to work in a client-server setup. In the simulated environment code, the client information is assumed to be available to the server before the experiment starts, but in a client-server setup, it needs to be shared by the client.
FjORD's client implementation attempts to send nested dictionaries and model weights, whereas Flower supports only string-to-scalar dictionaries, thus, the algorithm was not portable.

To confirm that the experimentation in \colext was not only successful but also produced the expected results, we executed the Moon~\cite{li2021moon} baseline and compared it with the FedProx~\cite{li2020fedprox} algorithm with 10 clients using the CIFAR-10 dataset as instructed in the README for the Moon baseline. The accuracy achieved for the final models was slightly lower than the values reported in the README, more specifically 0.1\% and 1.4\% lower for Moon and FedProx, respectively. This discrepancy can be explained by the alteration of the random number generator sequences when switching from a simulated environment to a client-server setup.

In conclusion, the experiments conducted in \colext demonstrate that the FL code can be successfully deployed in the testbed environment with only minor potential variations in accuracy compared to the simulated environment.

\subsubsection{Porting algorithms to Android devices}

On Android devices, Flower's support limits us to use TFLite. One significant constraint of TFLite is its lack of support for stateful FL algorithms.
To confirm the support of different algorithms in \colext, we needed to port stateless FL algorithms to Android.

Given the limitations, we developed an application that supports stateless algorithms, and we implemented and tested the algorithms presented in \autoref{tab:android_ported_algorithms}. The algorithms were ported from the Flower code repository. FedYogi and FedAdam modify the adaptive momentum estimation (Adam) optimizer to stabilize the learning process and improve convergence in heterogeneous federated settings. FedAdagrad adapts the Adagrad optimizer for FL by using per-parameter learning rates that adjust based on the history of gradients, accommodating non-IID data distributions. FedOpt serves as a generalized framework for federated optimization, encompassing various adaptive optimizers providing flexibility and robustness.

All of the stateless algorithms only require changes on the server side. Currently, the implementations in the Flower repository for all algorithms support serialization designed for Python implementations. To make the algorithms work with Android, we had to make certain changes to the code. The initial code from Flower relies on NumPy for the serialization of weights transferred to the clients and back. As we are running these clients on Android, we do not have access to NumPy, so we harnessed the existing implementation of a suitable serialization that is present in the FedAvgAndroid code from the Flower repository.  
Furthermore, the above algorithms require a random starting model. We added a mechanism similar to what FedAvg uses, whereby if there is no initial model present at the server, such a model is pulled from one randomly chosen client and broadcast to the others.

\begin{table}[tp]
    \centering
    \caption{FL algorithms tested on \colext smartphones.}    \label{tab:android_ported_algorithms}
    \begin{tabular}{|l|c|}\hline
        Algorithm &  Deployed\\\hline
        FedAvg~\cite{mcmahan2017fedavg}  &  \checkmark \\ \hline
        FedAdam~\cite{Wu2022-hm} & \checkmark \\ \hline
        FedYogi~\cite{Wu2022-hm}  & \checkmark \\ \hline
        FedAdagrad~\cite{Wu2022-hm}  & \checkmark \\ \hline
        FedOpt~\cite{asad2020fedopt}  & \checkmark \\ \hline
    \end{tabular}
\end{table}

\section{Profiling with \colext}

Heterogeneity of both SBC and smartphone devices included in our testbed, support for a broad span of algorithms and datasets, together with a range of performance metrics produced by the framework, ensure that \colext provides a rich experimentation space for FL research In this section, through a small number of use cases, we show how \colext experiments can be used to improve our understanding of FL.

\subsection{Revealing Accuracy vs. Resource Usage Trade-off}

\begin{figure}[tp]
    \centering
    \includegraphics[width=\linewidth]{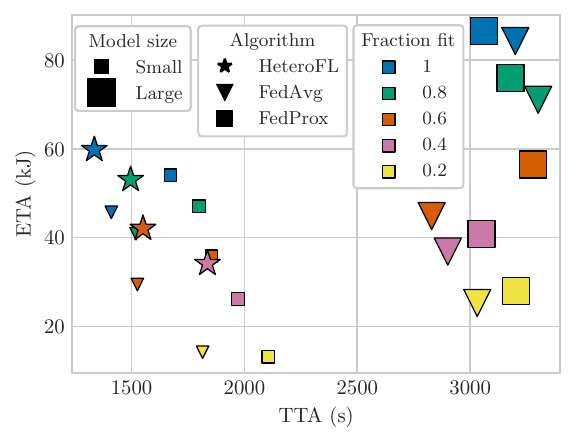}
    \caption{TTA and ETA consumed to reach $64\%$ accuracy on CIFAR-10 with $20$ clients and Dirichlet distribution ($\alpha = 1.0$). Missing configurations failed to reach the target accuracy for three consecutive evaluation rounds.}
    \label{fig:tta_vs_eta}
\end{figure}

The final model inference accuracy is the most reported metric in FL research. Yet, the accuracy is seldom juxtaposed against the resources needed to achieve it. In \autoref{fig:eta_vs_max_acc} of \autoref{sec:introduction}, we reveal that substantial energy costs can be incurred to achieve a very modest gain in accuracy. We now expand this investigation and assess how both the time and the energy vary as we aim to achieve a certain inference accuracy through different FL algorithms and settings.

We focus on the SBC clients in our testbed and assign both a training and a validation dataset sampled from the same distribution to each client, whereas distributions differ among the clients (i.e., non-IID data) according to the Dirichlet distribution with parameter $\alpha=1.0$. We employ FL with different algorithms (FedAvg, FedProx, and HeteroFL), using different model sizes (\emph{Small} -- 35k and \emph{Large} -- 380k parameters) and a different number of clients harnessed in each round. After each training round, each client evaluates the global model on its validation dataset and reports the resulting accuracy to the server. The experiment continues until the average validation accuracy across all clients achieves a pre-defined target value for three consecutive evaluation rounds.

We are interested in resources spent for training, thus we measure the wall clock (i.e., time-to-accuracy -- TTA) and the energy (i.e., energy-to-accuracy -- ETA) required to reach the target accuracy. To determine the energy spent on training, we first measure the average idle power consumption of the devices and subtract this ``average idle power'' from the power measurements during training. For instance, our experiments indicate that Jetson XavierNX consumes $2.9W$ (Watt) at rest. Hence, when such a device consumes, on average, $5W$ during the experiment, we consider that $2.1W$ is the ``average active power'' used for training. We then compute the energy used for training by multiplying the ``average active power'' by the time when the device is actively learning.

In \autoref{fig:tta_vs_eta}, we compare TTA and ETA for the three algorithms when the target accuracy is set to 64\%. We observe that using the \emph{Large} model (denoted with larger symbols) significantly increases the training time. A close look reveals that the cause for this lies in stragglers, as the server waits until it receives the updated models from all the sampled clients. Note that the FedAvg (square symbols) and the FedProx (inverse triangle symbols) algorithms require all the clients to train the same model architecture, while the HeteroFL algorithm (star symbols) goes beyond this limitation, allowing clients to train a model with a size proportional to their computational power. Consequently, the TTA for HeteroFL is comparable to the algorithms in which all clients train the \emph{Small} model (denoted with smaller symbols). For the cluster of points on the left of the figure, which reflect the training of the \emph{Small} model with the FedAvg and FedProx algorithms and heterogeneous model sizes with HeteroFL, we see a trend: increasing the percentage of clients sampled for training in each training round (``fraction fit'') decreases the TTA, as each update to the server model is obtained with more data. However, this causes an increase in ETA as more devices are used for training.

\subsection{Revealing (in)Efficiency of On-device Training}
\begin{figure}[tp]
    \centering
    \includegraphics[width=0.95\linewidth]{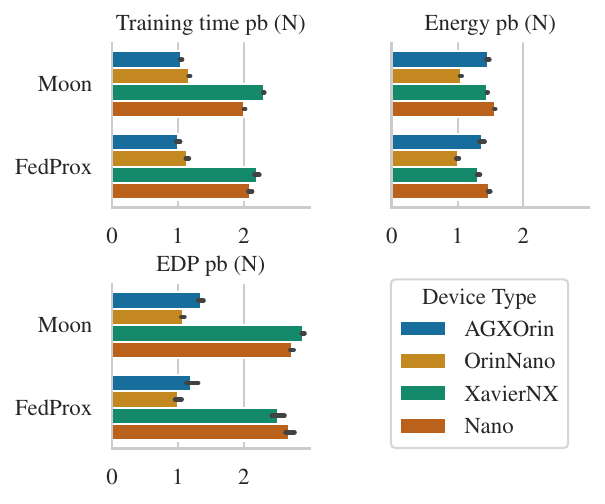}
    \caption{Comparing per batch (pb) efficiency of four NVIDIA Jetson models using the Moon and FedProx algorithms. All metrics have been Normalized (N) to the smallest value. The plotted data represents averages over 100 round samples, with error bars indicating the 95th percentiles.}
    \label{fig:moon_fed_prox_pb.pdf}
\end{figure}

\colext can also be used to assess the efficiency of different device types when handling the same workload.

We focus on the Moon and FedProx implementations that we experimented with in \autoref{porting_baselines}. To profile efficiency, we measure the time and the energy required to process one batch of data, calculated by dividing the measured time and energy required to finish a training round by the number of batches in the round. We deploy and run the algorithms on GPU-enabled SBCs in our testbed.

The results, shown in \autoref{fig:moon_fed_prox_pb.pdf} (top-left), reveal that the fastest device is, unsurprisingly, AGXOrin, which is also advertised as the most powerful SBC in our setup. However, for both algorithms, AGXOrin is not significantly faster than OrinNano, yet it uses noticeably more energy (top-right plot). Thus, we ask: How can we quantify whether the decrease in training time justifies the additional energy cost?

This led us to consider another metric common in digital electronics, the \emph{Energy Delay Product} (EDP). This metric multiplies the time taken to complete a task with the energy required for the task completion. Thus, a (preferred) low EDP indicates that a solution is both fast and energy-efficient. In \autoref{fig:moon_fed_prox_pb.pdf} (bottom-left), we observe that, according to EDP, OrinNano is the most efficient device in both algorithms. In other words, considering our subset of devices, if we only use OrinNano devices for this algorithm, we would minimize the time taken to train a batch for a given energy budget and vice-versa. The data also shows that XavierNX appears to be the slowest device, despite being a more recent model and having better hardware compared to the oldest model, the Jetson Nano (\autoref{tab:dev_spec} for device specifications). However, when FedProx is employed, despite XavierNX being the slowest device, it is still more efficient than a faster device, as is shown by the lower EDP value compared to Jetson Nano.

Note that our comparisons were done without fine-tuning the boards (e.g., the CPU operating frequency, etc.) for optimal efficiency;\footnote{Users can update the power mode and CPU frequency of the devices; however, currently this cannot be done via the \colext configuration file.} it is possible that running AGXOrin in a lower power mode could make it more efficient than OrinNano. Nevertheless, these examples demonstrate how EDP helps identify which devices are more efficient for specific algorithms and how device efficiency changes across algorithms. Another interesting use of EDP is to evaluate algorithm modifications to determine efficiency improvements.

With these insights, \colext provides users with a valuable tool to compare device efficiency, enabling them to observe and address potential performance issues with FL on real-world devices.

\subsection{Revealing Algorithm Implementation Issues}

\begin{figure}[tp]
    \centering
    \includegraphics[width=0.95\linewidth]{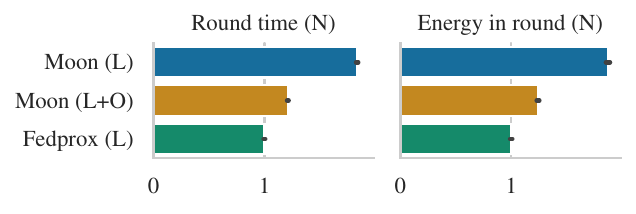}
    \caption{Normalized (N) round level metrics for Moon and FedProx using an 11.6M parameter model (L), with the addition of a small optimization (O) to the original Moon baseline code. Results show average round data statistics over 30 rounds, with error bars indicating the 95th percentiles.}
    \label{fig:cmp_algorithms_high_level.pdf}
\end{figure}

The testbed also allows us to analyze the algorithms' performance at the FL round level.  We once again turn back to the SBC-based experiment we performed with FedProx and Moon in the previous section. The algorithms are rather similar, yet we expect that results will reveal the cost of extra forward passes conducted by Moon -- compared to FedProx, which completes one forward pass, Moon performs three (one for the client model, two for the server model) in each round. However, the results of the experimentation (depicted in \autoref{fig:moon_fed_prox_pb.pdf} left) show that both algorithms take a very similar amount of time per batch (and, consequently, per round). To further unpack the issue, instead of using the \emph{Small} model, which might prevent the expected differences from being noticed, we tasked the algorithms with training the \emph{Large} model. This larger model exceeds the 4 GB memory of the Jetson Nano devices, so this experiment was only done with other Jetson device types. The results, shown in \autoref{fig:cmp_algorithms_high_level.pdf} (with L indicating the \emph{Large} model training), indicate that the difference in time between the two algorithms is obvious, if not unexpectedly large.

Surprised by such a large difference in the execution time (Moon $1.75 \times$ slower), we further inspected the code and identified unnecessary data movements from the GPU to the CPU memory in the Moon codebase. By addressing this and optimizing the code (denoted with L+O in \autoref{fig:cmp_algorithms_high_level.pdf}), the results finally get in line with the theoretical expectations regarding the two algorithms.

This experiment highlights how code optimizations and model size modifications can drastically change how two algorithms compare in terms of the training time, making their implementation and real-world performance analysis critical.

\subsection{Identifying Causes of Stragglers in Real-world Mobiles}

\begin{figure}[t!]
    \centering
    \begin{subfigure}[b]{0.5\textwidth}
        \centering
        \includegraphics[width=\linewidth]{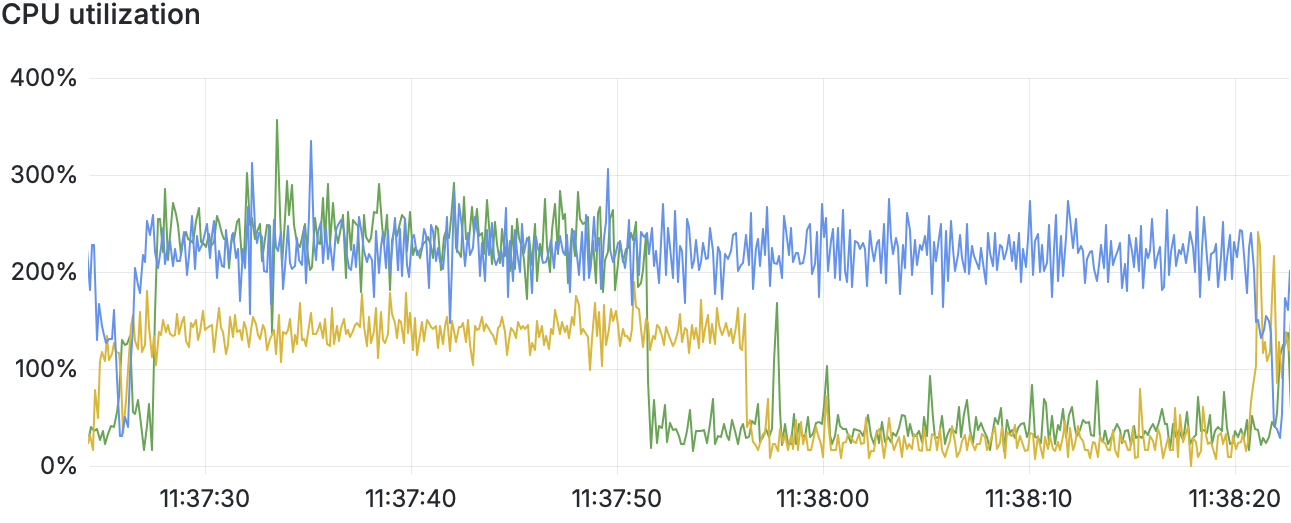}
    \end{subfigure}

    \begin{subfigure}[b]{0.5\textwidth}
        \centering
        \includegraphics[width=\linewidth]{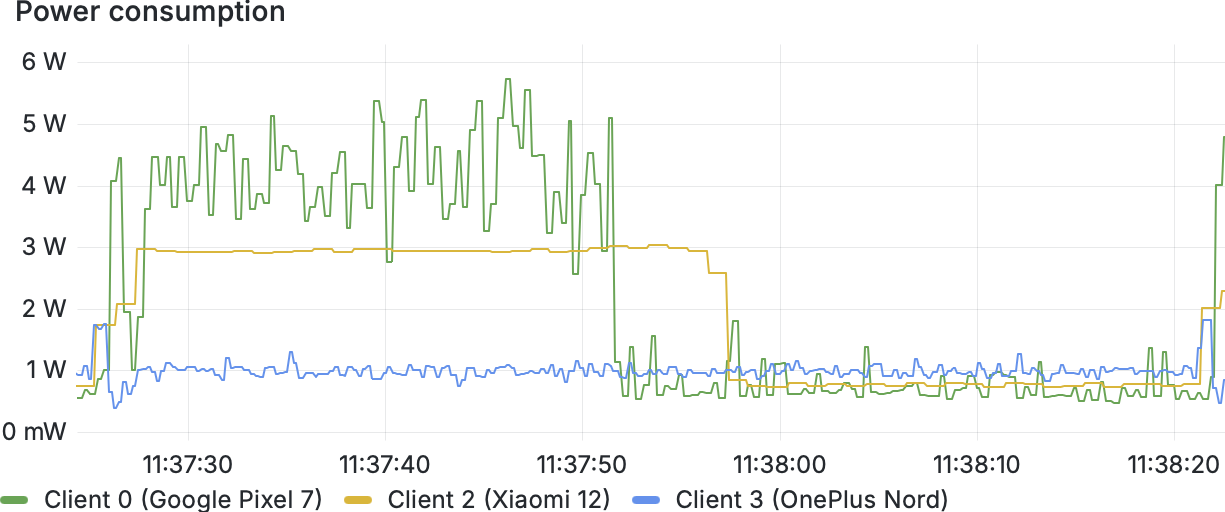}
    \end{subfigure}
    \caption{CPU utilization and power consumption on heterogeneous mobile hardware. Data collected over 1 round of training on CIFAR-10 dataset.}
    \label{fig:AndroidCPUProblem}
\end{figure}

\begin{figure}[t!]
    \centering
    \includegraphics[width=\linewidth]{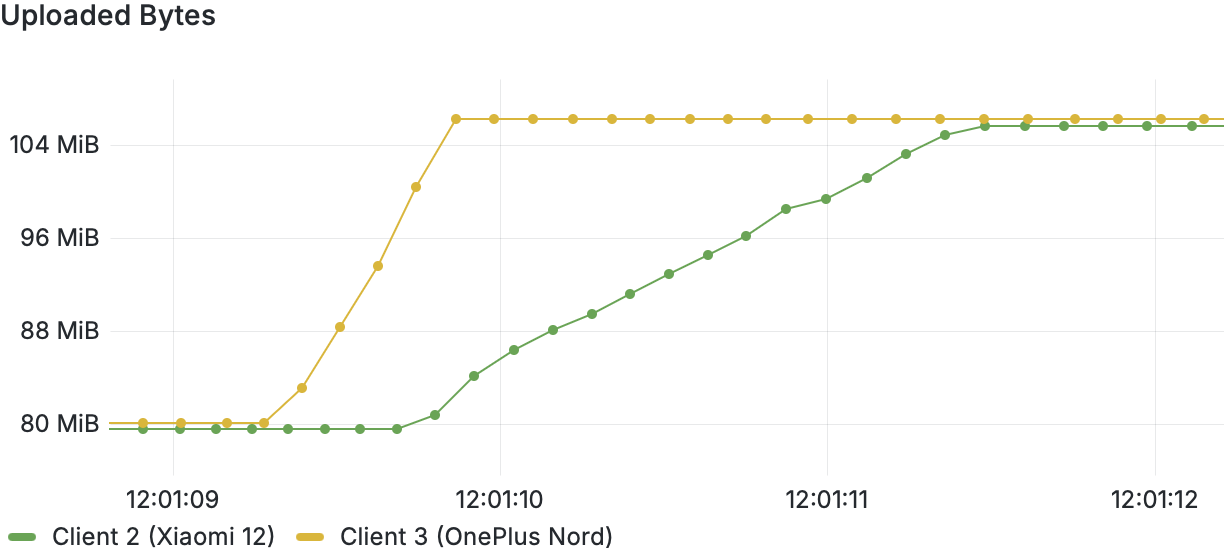}
    \caption{Uploaded bytes over time after a round of training on FEMNIST dataset~\cite{EMNIST} using CNN model.}
    \label{fig:AndroidUploadProblem}
\end{figure}

We perform extensive experimentation with Android devices in \colext and identify two main reasons for devices becoming stragglers, both due to hardware heterogeneity:

    \noindent \textbf{Compute speed.} Hardware heterogeneity introduces varying compute capabilities among devices, which is particularly noticeable during model training. The top graph of \autoref{fig:AndroidCPUProblem} shows the CPU utilization and power consumption during one round of training on three devices: Google Pixel 7, Xiaomi 12, and OnePlus Nord 2 5G. The start and end of local training can be identified by shifts in CPU utilization, from near idle (close to 0\%) to high utilization (over 100\% -- indicating that multiple CPU cores are active) and then back to idle. 
    We observe that OnePlus Nord takes the longest to complete a training round, with almost 2.3$\times$ longer round time than the fastest participating device in this training (Google Pixel 7). In the bottom graph of \autoref{fig:AndroidCPUProblem}, significant power consumption spikes are evident for two devices, contrasting with the OnePlus as a result of it being severely underclocked. Despite having similar CPUs, the default underclocking of the CPU in OnePlus makes a substantial difference in a real-world deployment of FL. 

    \noindent \textbf{Data reception and transmission.} Even with similar computing capabilities, devices can vary significantly in other hardware components, notably the wireless communication module (WiFi/cellular chip). With larger models requiring the transfer of many weights, differences in data transmission speeds become more evident. \autoref{fig:AndroidUploadProblem} illustrates this, showing the Xiaomi 12, with the default transmission parameters, as a straggler, taking 3.2$\times$  longer transmission time for the same amount of data compared to the OnePlus 2T 5G.

\section{Limitations and Future Work}
\label{sec:discussion}

\noindent \textbf{Android \& SBC deployments.}
Deep learning has evolved separately on desktop and server environments, where it is often implemented in Python, harnessing PyTorch or Keras libraries, and on mobile devices, where deep learning is usually based on the TensorFlow Lite (TFLite) library and written in Kotlin or Java. Consequently, FL is also supported in different manners on the two groups of devices present in the \colext testbed -- SBCs and Android phones. While we rely on Flower, due to its support for both of these device groups, we still face the barrier of different serialization formats used by PyTorch and TFLite, which hampers FL over a group of clients where Android and SBC devices are intermixed. A potential solution could be a compatibility-providing mapping between the PyTorch and TFLite serialization formats, something that ONNX~\cite{onnx} promises to provide. However, our extensive experiments with such a mapping failed to produce a reliable solution. Alternatively, deploying TFLite on single-board computers (SBCs) could be considered. However, this approach is not widely adopted among FL researchers, and TFLite may not offer the same level of usability as PyTorch for certain applications. In future versions of \colext, we aim to address this issue and enable a combination of models trained with SBC and Android devices to be used within the same experiment.

\noindent \textbf{Network conditions.}
Network variability is a common property of edge environments where wireless networks are the norm. \colext currently does not support controlling network conditions, but preliminary work on that front has already been conducted. For SBCs, we are adding support for fixed latency and bandwidth settings on clients, which could be configured using the Linux \texttt{tc} tool. Due to the difficulty of configuring Android clients in the same manner, for smartphones, we plan on supporting network parameter modifications on the server side only.

\noindent \textbf{Concurrent users.}
Given the size of our device pool, 28 SBCs and 20 smartphones, allowing multiple users on each platform would restrict FL experimentation with just a handful of devices, defeating the purpose of \colext, which aims to support the investigation of FL over heterogeneous hardware. Due to this, access to \colext is currently limited to two users at a time for a set period (e.g., one week), with one user assigned to the Android portion, and the other user to the SBCs. Once the number of devices is significantly increased, a more scalable access mechanism will be introduced.

\section{Conclusion}
\label{sec:conclusion}

Reproducible and realistic experimentation is necessary for the future growth of distributed AI. Despite its research popularity, FL remains mostly confined to simulations, as the effort of deploying FL solutions over heterogeneous edge devices and collecting performance metrics remains insurmountable to most research groups. We presented \colext, an FL experimentation framework and testbed that allows an arbitrary FL algorithm to be run in a heterogeneous environment and assessed from various performance aspects. Our testbed already employs over 40 devices and collects a range of metrics, including inference accuracy, detailed energy usage, and CPU/GPU/memory utilization information. Nevertheless, we believe that by making \colext publicly available, we will support the organic growth of our testbed so that it addresses the up-to-date needs of FL researchers and practitioners.
\section*{Acknowledgments}
This publication is based upon work supported by the King Abdullah University of Science and Technology (KAUST) Office of Research Administration (ORA) under Award No. ORA-CRG2021-4699, and partly supported by the Slovenian Research Agency (research projects J2-3047 and P2-0098).
We are thankful to Abdullah Alamoudi, Suliman Alharbi, Rasheed Alhaddad for their helpful contributions to a first prototype of our system.
Finally, we wish to thank the reviewers and our shepherd for their thoughtful comments and suggestions that helped improving our paper.

\bibliographystyle{IEEEtran}
\bibliography{main}

\end{document}